\documentclass[preprint,12pt]{elsarticle}

\usepackage[margin=2.5cm]{geometry}
\usepackage{amssymb}
\usepackage{amsthm}
\usepackage{lineno}
\usepackage{xcolor}
\graphicspath{{plot/}}
\usepackage{subcaption}
\usepackage{tabularx}
\usepackage{array}
\usepackage{amsmath,bm}
\usepackage{multirow}
\usepackage{hyperref}
\usepackage{upgreek}

\newcommand{\Bm}{\mathbf{m}}
\newcommand{\R}{\mathbb{R}}
\newcommand{\Bg}{g}
\newcommand{\Bd}{\mathbf{d}}

\newcommand{\Bdhm}{\mathbf{d}_\text{hm}}
\newcommand{\Nr}{N_\text{r}}

\newcommand{\Bmtrue}{\mathbf{m}_\text{true}}
\newcommand{\Bdobs}{\mathbf{d}_\text{obs}}

\newcommand{\CD}{C_{\text{D}}}
\newcommand{\Bxi}{\bm{\upxi}}
\newcommand{\Nl}{N_\text{l}}
\newcommand{\Bdhat}{\tilde{\mathbf{d}}}
\newcommand{\Na}{N_\text{a}}
\newcommand{\Bi}{\mathbf{i}}
\newcommand{\Bf}{\mathbf{f}}
\newcommand{\Bo}{\mathbf{o}}
\newcommand{\BW}{\mathbf{W}}

\newcommand{\Bc}{\mathbf{C}}
\newcommand{\Bx}{\mathbf{X}}
\newcommand{\Bh}{\mathbf{H}}
\newcommand{\Bchat}{\Tilde{\mathbf{C}}}

\newcommand{\Bb}{\mathbf{b}}

\begin{document}

\begin{frontmatter}

\title{History Matching for Geological Carbon Storage using Data-Space Inversion with Spatio-Temporal Data Parameterization}
\author[1]{Su Jiang\corref{cor1}}
\author[1]{Louis J.~Durlofsky}

\address[1]{Department of Energy Science and Engineering, Stanford University, Stanford, CA 94305, USA}

\begin{abstract}
History matching based on monitoring data will enable uncertainty reduction, and thus improved aquifer management, in industrial-scale carbon storage operations. In traditional model-based data assimilation, geomodel parameters are modified to force agreement between flow simulation results and observations. In data-space inversion (DSI), history-matched quantities of interest, e.g., posterior pressure and saturation fields conditioned to observations, are inferred directly, without constructing posterior geomodels. This is accomplished efficiently using a set of $O(1000)$ prior simulation results, data parameterization, and posterior sampling within a Bayesian setting. In this study, we develop and implement (in DSI) a deep-learning-based parameterization to represent spatio-temporal pressure and CO$_2$ saturation fields at a set of time steps. The new parameterization uses an adversarial autoencoder (AAE) for dimension reduction and a convolutional long short-term memory (convLSTM) network to represent the spatial distribution and temporal evolution of the pressure and saturation fields. This parameterization is used with an ensemble smoother with multiple data assimilation (ESMDA) in the DSI framework to enable posterior predictions. A realistic 3D system characterized by prior geological realizations drawn from a range of geological scenarios is considered. A local grid refinement procedure is introduced to estimate the error covariance term that appears in the history matching formulation. Extensive history matching results are presented for various quantities, for multiple synthetic true models, using the new DSI framework. Substantial uncertainty reduction in posterior pressure and saturation fields is achieved in all cases. The framework is also applied to efficiently provide posterior predictions for a range of error covariance specifications. Such an assessment would be very expensive using a traditional model-based approach.

\end{abstract}


\end{frontmatter}

\section{Introduction}
Extensive monitoring and data collection will be required in large-scale geological carbon storage operations. When used in combination with effective data assimilation (or history matching) methodologies, these monitoring data can provide substantial uncertainty reduction in performance predictions. Traditional model-based data assimilation methods calibrate model parameters, such as grid-block permeability and porosity values, by minimizing the mismatch between observed data and model predictions. Because these predictions require large numbers of flow simulations this approach can be very time-consuming, especially if a wide range of geological scenarios is considered. 

In this study we extend and apply an alternative approach, referred to as data-space inversion (DSI), for carbon storage problems. DSI methods do not provide posterior geomodels, which represents a disadvantage in some settings, and they require a preprocessing step involving flow simulation for a substantial number (e.g., 1000--2000) of prior geological models. However, once these prior flow simulations are performed, DSI can directly generate posterior (history matched) predictions for quantities of interest (QoI), such as pressure and CO$_2$ saturation fields at a set of time steps, conditioned to observed data. These predictions can be generated very quickly, without the need for any additional flow simulations. This capability enables DSI to efficiently assess the impact of different types and amounts of measured data, different monitoring well designs, varying measurement and model errors, etc. With traditional model-based history matching, each of these assessments could require thousands of additional flow simulations. Our specific goals here are to develop a deep-learning-based data parameterization method in DSI to capture the spatio-temporal evolution of pressure and CO$_2$ saturation fields, and to apply these new treatments to cases in which the geological scenario (characterized by a set of metaparameters) is itself uncertain. 

Model-based data assimilation methods have been widely applied in geological carbon storage. Our discussion here will be brief as we will not be using these traditional approaches in this study. Model-based methods have been used with multiple types of monitoring data, including flow and geophysical data, to predict pressure, saturation and geomechanical effects. \citet{cameron2016use}, for example, developed a procedure that applied pressure data to identify leakage (location and volume) through the caprock. \citet{gunning2020bayesian} used pressure tests and nonlinear adjoint theory to quantify the location of CO$_2$ plumes. \citet{espinet2013estimation} incorporated saturation data, which improved forecasts for plume location. \citet{xiao2022model} developed a model-reduced adjoint inversion to predict 2D CO$_2$ saturation fields using saturation measurements. \citet{chen2020reducing} assimilated both pressure and saturation data and proposed an inflated ensemble-based treatment to reduce uncertainty in plume metrics. \citet{ma2019dynamic} showed that the inclusion of time-lapse seismic data led to improvement in the calibration of aquifer properties. \citet{jahandideh2021inference} used microseismic observations to predict flow properties and geomechanical responses. 

A wide variety of deep-learning-based surrogate models have been developed to reduce the computational costs associated with the flow simulations required for history matching. \citet{tang2021deep_co2} developed a recurrent R-U-Net architecture to predict the pressure, saturation and surface displacement in geological carbon storage settings. \citet{wen2022u} developed a U-Net enhanced Fourier neural operator (FNO) to approximate flow responses for single well injection scenarios. \citet{yan2022robust} applied FNO to a realistic example in a 3D heterogeneous saline aquifer. \citet{jiang2023fourier} developed Fourier-enhanced multiple-input neural operators for carbon storage to reduce the number of parameters and the training time. Although many efficient surrogate models have been developed, challenges still exist in the treatment of general cases involving, e.g., multiple (deviated or horizontal) injectors, fractures and faults, unstructured grids, etc. In addition, in some studies, large numbers of training runs (20,0000 in \cite{wen2021towards}) and extensive training times are required.

The DSI framework used here was originally developed for oil field applications by \citet{Sun2017} and~\citet{SunCG}. Data variables (QoI) consisted of time series of water and oil flow rates at injection and production wells. Subsequent studies, again in oil field settings, extended these initial DSI formulations. \citet{jiang2019data}, for example, enhanced the method to treat varying well pressures, which enabled production optimization to be performed. \citet{lima2019data} evaluated a number of treatments and incorporated more efficient posterior sampling into DSI. \citet{hui2023data} applied DSI to forecast unconventional reservoir performance in real-world cases. The original DSI formulation was also extended to CO$_2$ storage settings \cite{sun2019data}, where it was used to predict CO$_2$ saturation for the top layer of the storage aquifer at the end of the simulation time frame. We note that DSI shares similarities with other direct forecasting methods, such as those described in \cite{scheidt2015prediction, satija2015direct, satija2017direct, hermans2016direct, jeong2018learning}. The relationship between direct forecasting methods and DSI is discussed in \cite{Sun2017}.

In the context of model-based history matching, deep-learning parameterizations have been shown to be very useful for representing realistic geomodels (e.g., 3D channelized systems) with a relatively small set of latent variables. See, e.g., \cite{laloy2018training, laloy2019gradient, liu20213d, han2022characterization} for methodologies and discussion. In the context of DSI, parameterization is similarly important. Principal component analysis (PCA) with histogram transformation has been used in several DSI studies \cite{SunCG, sun2019data, hui2023data}. This approach is often effective, but can yield inaccurate results for correlations between QoI and for derived quantities because histogram transformation preserves only marginal distributions. To address this issue, \citet{jiang2021data} developed a recurrent autoencoder to parameterize flow rate (time-series) data. This approach, which was shown to preserve important correlations in QoI in synthetic cases, was subsequently applied to a realistic fractured reservoir system \cite{jiang_frontiers}. \citet{xiao2023deep} introduced a convolutional autoencoder to parameterize CO$_2$ saturation data. The temporal dynamics of the saturation measurements from multiple monitors were reorganized as a spatial field.

Dimension reduction in terms of latent variables has been used to represent spatial-temporal fields in many application areas. \citet{kang2021integrating}, for example, applied a convolutional variational autoencoder (CVAE) to parameterize nonaqueous phase liquid saturation. The procedure was then used for the monitoring of source zones. \citet{kao2021fusing} developed a stacked autoencoder to represent flood features. They integrated recurrent neural networks to predict regional inundation maps. \citet{shamekh2023implicit} applied an autoencoder to parameterize high-resolution spatial distributions of water, which were coupled with deep neural networks to predict precipitation. To our knowledge, spatio-temporal data parameterizations have not yet been applied in carbon storage settings.  

Along these lines, a key goal of this work is the development of a spatio-temporal data parameterization to capture the spatial features and temporal dynamics of pressure and CO$_2$ saturation fields in geological carbon storage systems. This representation is then incorporated into the DSI framework to enable accurate history matched pressure and saturation predictions. We introduce an adversarial autoencoder (AAE) to generate data variables (pressure and saturation fields that follow prior statistics) from normally distributed latent variables. Convolutional long short-term memory (convLSTM) is incorporated within the AAE to treat time evolution. The AAE-based parameterization is used with an ensemble smoother with multiple data assimilation (ESMDA) to generate posterior DSI predictions. The AAE parameterization improves upon PCA with histogram transformation as it preserves correlations in the data variables, and it is compatible with ESMDA because it ensures the latent variables follow a Gaussian distribution.  

The new DSI treatment is applied to 3D heterogeneous geomodels based on a storage system now under development in the Mount Simon formation in Illinois, USA \cite{okwen2022storage}. The prior geomodels are realizations sampled from a range of geological scenarios characterized by a set of metaparameters. The flow scenario entails 4~Mt/year of CO$_2$ injection through four horizontal wells for a 20-year period. Measurement data include pressure and saturation from eight vertical monitoring wells. A local grid refinement assessment is applied to quantify model-resolution error, which is an important quantity in history matching formulations. DSI predictions for a range of QoI are presented. The impact of varying levels of measurement and model error on predictions is quantified. Model performance is evaluated for multiple (synthetic) true models.

This paper proceeds as follows. In Section~\ref{sec:method}, we review the DSI method with data parameterization. The deep-learning-based parameterization (AAE with convLSTM) developed to represent the spatio-temporal pressure and saturation fields is then described. The geomodels and problem setup are presented in Section~\ref{sec:prior}. The ability of the new parameterization to reconstruct data realizations and to generate new realizations consistent with prior statistics is demonstrated. In Section~\ref{sec:post}, we first estimate model resolution error using a local grid refinement procedure. Extensive posterior DSI results are then presented. Conclusions and suggestions for future work are provided in Section~\ref{sec:conclusion}. 
\section{Data-space inversion with data parameterization}\label{sec:method}

In this section, we extend the basic data-space inversion (DSI) method, presented in \citep{Sun2017,SunCG,jiang2021data}, to enable the prediction of pressure and CO$_2$ saturation fields at a number of time steps. We then introduce a new deep-learning-based parameterization for this type of spatio-temporal data. The parameterization entails the use of an adversarial autoencoder (AAE) and a convolutional long short-term memory network (convLSTM). Finally, we discuss the combination of the new AAE parameterization with an ensemble smoother with multiple data assimilation (ESMDA). 

\subsection{DSI formulation with spatio-temporal data}
In the DSI procedure, posterior forecasts for quantities of interest (QoI) conditioned to observations are generated directly, without calibrating the geomodel parameters. The correlations between data in the historical period and data in the forecast period are exploited to enable this direct forecasting. An ensemble of prior model realizations ($\sim$1000) are simulated, through both the historical and prediction periods, to capture these correlations. A Bayesian framework is then applied to generate the posterior distributions of QoI in the forecasting period. 

We denote the prior geological realizations as $\Bm_i \in \R^{n_b \times 1} $, $i = 1, \ldots, \Nr$, where $n_b$ represents the number of grid blocks and $\Nr$ represents the number of prior realizations. These geological models define all necessary quantities, e.g., porosity and permeability, in each grid block. In cases where porosity and permeability are independent, or multiple permeability components are defined, we have $\Bm_i \in \R^{n_p n_b \times 1} $, $i = 1, \ldots, \Nr$, where $n_p$ is the number of geological parameters per grid block. The realizations are generated using geostatistical modeling software, and they may be drawn from either a single geological scenario or multiple scenarios. In this work we consider multiple scenarios characterized by different multi-Gaussian correlation structures.

The pressure and CO$_2$ saturation fields at a particular time step are denoted $\mathbf{x}^{\text{sim}}_P \in \R^{n_b \times 1}$ and $\mathbf{x}^{\text{sim}}_S \in \R^{n_b \times 1}$. Prior states are generated from forward simulation, denoted as $\Bg$, i.e., 
\begin{equation}\label{eq:forward}
    [(\mathbf{x}^{\text{sim}}_P)_{i, k}, (\mathbf{x}^{\text{sim}}_S)_{i, k}] = \Bg(\Bm_i),
\end{equation}
for $i = 1, \ldots, \Nr$, and $k = 1, \ldots, n_{ts}$, where $k$ denotes time step and $n_{ts}$ represents the number of simulation time steps. The simulation data corresponding to the historical period are denoted $\Bdhm^{\text{sim}} \in \R^{N_{\text{hm}} \times 1}$, where $N_{\text{hm}} = 2\times n_b \times n_{th}$, with $n_{th}$ the number of time steps in the historical period. The data vector $\Bdhm^{\text{sim}}$, for $i=1, \dots, N_r$, is written,
\begin{equation}
    (\Bdhm^{\text{sim}})_i = [(\mathbf{x}^{\text{sim}}_P)_{i, 1}, (\mathbf{x}^{\text{sim}}_S)_{i, 1}, \ldots, (\mathbf{x}^{\text{sim}}_P)_{i, n_{th}}, (\mathbf{x}^{\text{sim}}_S)_{i, n_{th}}].
\end{equation}
The full data vector $\Bd^{\text{sim}}_{\text{full}} \in \R^{N_{\text{full}} \times 1}$ contains the pressure and saturation fields over the full simulation time frame, i.e., for both the historical and prediction periods. This data vector, denoted $\Bd^{\text{sim}}_{\text{full}}$, $i=1, \dots, N_r$, is expressed as

\begin{equation}
(\Bd^{\text{sim}}_{\text{full}})_i = [(\mathbf{x}^{\text{sim}}_P)_{i, 1}, (\mathbf{x}^{\text{sim}}_S)_{i, 1}, \ldots, (\mathbf{x}^{\text{sim}}_P)_{i, n_{th}}, (\mathbf{x}^{\text{sim}}_S)_{i, n_{th}}, \ldots, (\mathbf{x}^{\text{sim}}_P)_{i, n_{t}}, (\mathbf{x}^{\text{sim}}_S)_{i, n_{t}}], 
\end{equation}
where $n_{t}$ is the number of time steps considered for the full period, and $N_{\text{full}} = 2\times n_b \times n_{t}$. Note that $n_{t}$ is generally different than the number of simulation time steps $n_{ts}$. This is because, by using a representative set of time steps with $n_t \ll n_{ts}$, substantial computational savings can be achieved.

The observed data $\Bdobs^{\text{tot}} \in \R^{N_{\text{obs}} \times 1}$ considered in this work include pressure and saturation measurements from multiple monitoring wells during the historical period. Here $N_{\text{obs}}$ is the number of observed data. 
For the synthetic cases considered in this study, we do not have actual measurements. Instead, we generate observed data $\Bdobs^{\text{tot}}$ from a `true' model $\Bmtrue$, which is a randomly selected realization not included in the prior ensemble. The observed data $\Bdobs^{\text{tot}}$ are obtained by adding random error to `true' data $\Bd_{\text{true}} \in \R^{N_{\text{obs}} \times 1}$, with the true data generated via simulation of $\Bm_{\rm true}$. The observed data $\Bdobs^{\text{tot}}$ (which include error) are thus expressed as 
\begin{equation}\label{eq:dobs}
    \Bdobs^{\text{tot}} = \Bd_{\text{true}} + \mathbf{e},
\end{equation}
where $\mathbf{e} \in \R^{N_{\text{obs}} \times 1}$ denotes the error in the observations. 

In many data assimilation procedures, only measurement errors, sampled from a Gaussian distribution $N(\mathbf{0}, \CD)$, where $\CD$ is the error covariance matrix, are included in $\mathbf{e}$. The instrumentation used in practice, however, measures pressure and saturation on a scale of cm to m. Even high-fidelity simulation models, with tens of millions of cells, do not achieve this level of resolution. Therefore, in this work, the error covariance will include both measurement error and model-resolution error, with the latter accounting for scale discrepancies. The detailed error setup will be described in Section~\ref{sec:LGR}. 

The true data $\Bd_{\text{true}}$ for the synthetic case include the pressure and saturation data from all monitoring wells in the historical period. These data are represented as 
\begin{equation}
\Bd_{\text{true}} = [\Bd_{\text{true}}^1, \Bd_{\text{true}}^2, \ldots, \Bd_{\text{true}}^{n_{mw}}], 
\end{equation}
where $\Bd_{\text{true}}^l \in \R^{N_{\text{obs}}^l \times 1}$ denotes the measurement data at each monitoring well $l$, for $l = 1, \ldots, {n_{mw}}$, and $n_{mw}$ is the number of monitoring wells. The true data vector $\Bd_{\text{true}}^l$ from each monitoring well includes both pressure and saturation data and is given by 
\begin{equation}
    \Bd_\text{true}^l = [(\mathbf{x}_P)^l_{\text{true}, 1}, (\mathbf{x}_S)^l_{\text{true}, 1}, \ldots, (\mathbf{x}_P)^l_{\text{true}, n_{th}}, (\mathbf{x}_S)^l_{\text{true}, n_{th}}], 
\end{equation}
where $(\mathbf{x}_P)^l_{\text{true}, t}$, for $k = 1, \ldots, n_{th}$, represents the pressure data for the true model, at monitoring well $l$ at simulation time step $k$, and $(\mathbf{x}_S)^l_{\text{true}, t}$, for $t = 1, \ldots, n_{th}$, denotes the true saturation data at monitoring well $l$ and time step $t$. 

The true pressure data vector $(\mathbf{x}_P)^l_{\text{true}, t}$ can be expressed as $(\mathbf{x}_P)^l_{\text{true}, t} = E^l_P (\mathbf{x}_P)_{\text{true}, t},$ and the true saturation data vector as $(\mathbf{x}_S)^l_{\text{true}, t} = E^l_S (\mathbf{x}_S)_{\text{true}, t}$. Here $E^l_P \in \R^{(n_w)_P^l \times n_b}$ and $E^l_S \in \R^{(n_w)_S^l \times n_b}$ are extraction matrices, containing ones and zeros, which simply extract the pressure and saturation observations corresponding to the historical data measured at each monitoring well $l$. The quantities $(n_w)_P^l$ and $(n_w)_S^l$ are the number of pressure and saturation observations in monitoring well $l$. The total number of observations at monitoring well $l$ at a particular time step is $N_{\text{obs}}^l = (n_w)_P^l + (n_w)_S^l$, and the total number of observations over all monitoring wells and all times is $N_{\text{obs}} = N_{\text{obs}}^l \times n_{th}$.

The Bayesian framework is then applied in the data space to generate posterior distributions of data variables $\Bd^{\text{sim}}$. The posterior probability density function (PDF) of data variables $\Bd^{\text{sim}}$, conditioned to pressure and saturation observations $\Bdobs^\text{tot}$, is expressed as 
\begin{equation}\label{eq:post_pdf_{dec}}
    p(\Bd^{\text{sim}}|\Bdobs^\text{tot}) = \frac{p(\Bdobs^\text{tot} | \Bd^{\text{sim}})p(\Bd^{\text{sim}})}{p(\Bdobs^\text{tot})} \propto p(\Bdobs^\text{tot}|\Bd^{\text{sim}}) p(\Bd^{\text{sim}}),
\end{equation}
where $p(\Bdobs^\text{tot} | \Bd^{\text{sim}})$ denotes the conditional PDF of observed data $\Bdobs^\text{tot}$ given data variables $\Bd^{\text{sim}}$, and $p(\Bd^{\text{sim}})$ denotes the prior PDF of data variables $\Bd^{\text{sim}}$. 

\subsection{Spatio-temporal data parameterization with adversarial autoencoder (AAE)}
The data variables $\Bd^{\text{sim}}$ in DSI are generated from nonlinear multiphase flow simulations so they are, in general, high-dimensional non-Gaussian distributed. Both the prior and posterior distributions in Eq.~\ref{eq:post_pdf_{dec}} are, therefore, complicated to quantify. Data parameterization provides an effective way to reduce the dimension of $\Bd^{\text{sim}}$ and to represent the non-Gaussian distributions using approximately Gaussian-distributed low-dimensional latent variables $\Bxi \in \R^{\Nl \times 1}$. Here $\Nl$ is the dimension of the latent variables, with $\Nl \ll N_{\rm full}$. The mapping from latent vector $\Bxi$ to data vector $\Bd^{\text{sim}}$ is expressed as $\Bd^{\text{sim}} \approx \Bdhat^{\text{sim}} = f(\Bxi)$. Assuming prior samples of latent variables $\Bxi$ follow a Gaussian distribution, the posterior probability of $\Bxi$, conditioned to the observed data $\Bdobs^{\text{tot}}$, is given by 
\begin{equation}\label{eq:post_pdf_xi}
p(\Bxi|\Bdobs^{\text{tot}}) \propto \text{exp} \left(-\frac{1}{2}(Ef(\Bxi) - \Bdobs^{\text{tot}}) ^T\CD^{-1}(Ef(\Bxi)- \Bdobs^{\text{tot}}) -\frac{1}{2}\Bxi^T\Bxi \right), 
\end{equation}
where $E \in \R^{N_{\rm obs} \times N_{\rm full}}$ is the total extraction matrix, which is applied to collect the data in the historical period corresponding to the observations. 

\citet{SunCG} developed a parameterization method that combines principal component analysis (PCA) with histogram transformation (HT) to represent time series of well-rates in oil-water subsurface flow problems. The treatment was extended to geological carbon storage systems to parameterize 2D maps, specifically CO$_2$ saturation in a particular layer at a single time step~\cite{sun2019data}. PCA with HT parameterization is flexible but it preserves only the marginal distributions, not the correlations between the high-dimensional data variables.

To address this important limitation, a deep-learning-based recurrent autoencoder (RAE) parameterization was introduced in \cite{jiang2019data}. This procedure was shown to capture the high-dimensional distributions and correlations in multivariate flow-rate time series data. This approach is not, however, directly applicable for  spatio-temporal data, i.e., for the time-evolving pressure and saturation fields of interest here. This motivates the development of the new deep-learning-based spatio-temporal data parameterization treatment, involving an adversarial autoencoder (AAE) and a long-short-term memory convolutional (convLSTM) recurrent network, which we now describe.

\subsubsection{Convolutional long short-term memory (convLSTM)}
Long short-term memory (LSTM)~\cite{hochreiter1997long}, a variant of recurrent neural networks, was used to represent the temporal evolution of flow rates, in both the encoder and decoder, in the RAE architecture presented in \cite{jiang2019data}. Here, the convolutional LSTM (convLSTM) is applied in the autoencoder to represent spatial features and temporal dynamics in the (evolving) pressure and saturation states. The 3D convolutional neural networks (CNNs) maintain spatial information, while the LSTM captures temporal evolution. Stacked 3D convLSTM layers are applied in the encoder and decoder to provide accurate representations. The convLSTM is shown in Fig.~\ref{fig:aae}(a). We now describe the detailed architecture of this network.

The network comprises a chain structure of repeating 3D convLSTM cells with the same weights. At each time step $t$, for $t =  1, 2, \ldots, n_t$, the network cell acts on input $\Bx^t \in \R^{n_x \times n_y \times n_z \times n_c}$, the output state from the previous cell $\Bh^{t-1}$, and the cell state from the previous cell $\Bc^{t-1}$. Here, the input $\Bx^t$ includes the 3D spatial data of dimension $n_b = n_x \times n_y \times n_z$ from $n_c$ channels ($n_x$, $n_y$ and $n_z$ are the number of grid blocks in each direction). In the first layer of the encoder, the input is the time series of pressure and saturation fields. The number of channels is thus 2. Three gates, referred to as the forget gate $\Bf^t$, the input gate $\Bi^t$, and the output gate $\Bo^t$, control the memory flow in convLSTM networks. These are determined from the previous output state $\Bh^{t-1}$ and the input $\Bx^t$ through the following expressions
\begin{equation}
    \begin{split}
        &\Bf^t = \sigma(\BW_{f, H}\Bh^{t-1}+\BW_{f, X}\Bx^{t}+ \Bb_f), \\
        &\Bi^t = \sigma(\BW_{i, H}\Bh^{t-1}+\BW_{i, X}\Bx^{t} + \Bb_i), \\
        &\Bo^t = \sigma(\BW_{o, H}\Bh^{t-1}+\BW_{o, X}\Bx^{t} + \Bb_o), \\    
    \end{split}
\end{equation}
where the $\BW$ matrices and $\Bb$ vectors represent the convolutional kernel weights and bias terms (the weights are shared in the cells for each time step), and $\sigma(\cdot)$ is the sigmoid function, which operates component-wise on each variable. 

The forget gate $\Bf^t$ controls which information is retained (versus forgotten) from the previous cell state $\Bc^{t-1}$. The input gate $\Bi^t$ controls the information from the input $\Bx^t$, and the output gate $\Bo^t$ determines the output $\Bh^t$ from the current cell state $\Bc^t$. The new candidate cell state $\Bchat^t$ is given by 
\begin{equation}
    \Bchat^t = \text{tanh}(\BW_{c, H}\Bh^{t-1}+\BW_{c, X}\Bx^{t} + \Bb_c).
\end{equation}
The cell state $\Bc^{t}$ is then computed from the candidate state $\Bchat^t$ and the previous state $\Bc^{t-1}$ through application of
\begin{equation}
    \Bc^t = \Bf^t \circ \Bc^{t-1} + \Bi^t \circ \Bchat^t,
\end{equation}
where $\circ$ indicates the element-wise Hadamard product. The output state $\Bh^t$ 
is then given by 
\begin{equation}
    \Bh^t = \Bo^t \circ \text{tanh}(\Bc^t).
\end{equation}
The incorporation of 3D convLSTM in the autoencoder and adversarial autoencoder is discussed in the next section. 

\subsubsection{Adversarial autoencoder architecture} 
\begin{figure}[htbp!]
\centering
\begin{minipage}{.9\linewidth}\centering
\includegraphics[trim = 0 100 0 120, clip, width=\linewidth]{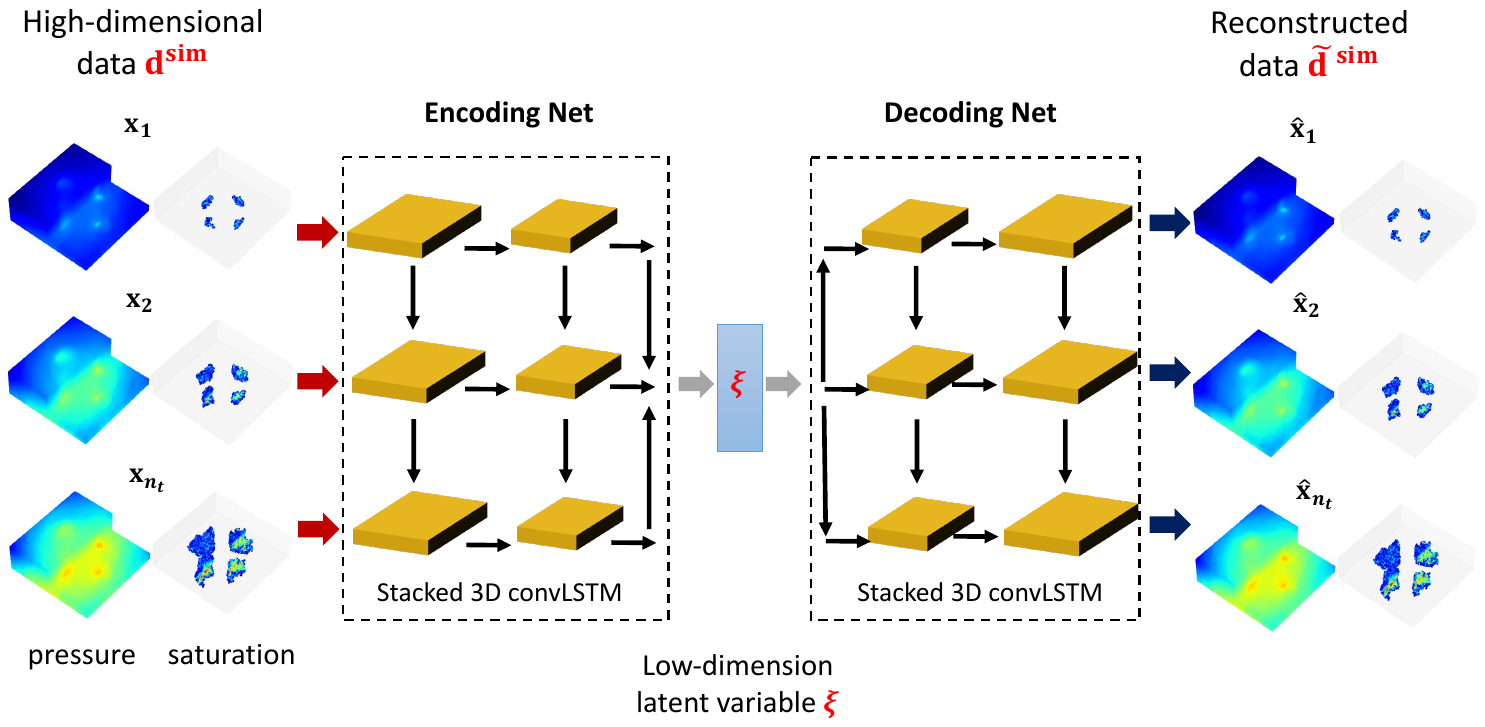}
\subcaption{Encoder and decoder with stacked convLSTM layers}
\end{minipage}
\centering
\begin{minipage}{.75\linewidth}\centering
\includegraphics[trim = 0 140 0 120, clip, width=\linewidth]{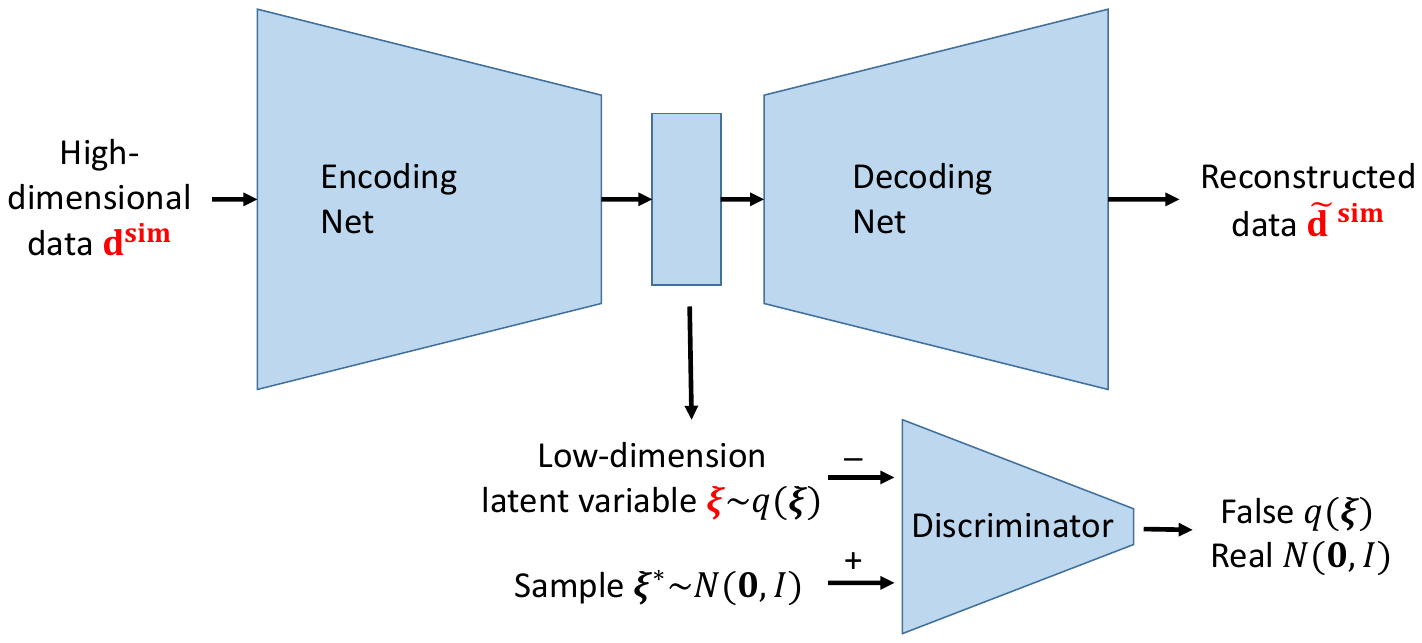}
\subcaption{Adversarial autoencoder architecture with autoencoder and discriminator}
\end{minipage}

\caption{Schematic of adversarial autoencoder (AAE) with stacked convLSTM layers (see Table~\ref{tab:aae} for detailed specifications).}\label{fig:aae}
\end{figure}

An autoencoder with 3D convLSTM is used in this work (Fig.~\ref{fig:aae}). Both the encoder and decoder are constructed from stacked convLSTM layers to improve the reconstruction accuracy. The encoder maps data variables $\Bd^{\text{sim}}$ to latent variables $\Bxi$, and the decoder maps $\Bxi$ back to the high-dimensional data vector. The encoding process is expressed as $\Bxi = f_{enc}(\Bd^{\text{sim}}; W_{enc})$, where $f_{enc}$ represents the encoder network and $W_{enc}$ denotes the weights in the encoder. The reconstructed high-dimensional data variables $\Bdhat^{\text{sim}}$ are generated via $\Bdhat^{\text{sim}} = f_{dec}(\Bxi; W_{dec})$, where $f_{dec}$ is the decoder network and $W_{dec}$ the weights in the decoder. 

High-dimensional statistics are (approximately) preserved through the nonlinear mapping in many autoencoder procedures. The latent variables are often close to Gaussian distributed, though they are not guaranteed to be strictly Gaussian. Thus the latent variables may not be in full compatibility with the Gaussian assumptions inherent in many posterior sampling methods, including the ESMDA procedure used in this work. 

For this reason, we introduce an adversarial autoencoder \cite{makhzani2015adversarial} with the ability to (1) reproduce/generate new data realizations that follow the prior statistics, and (2) provide Gaussian latent variables that are consistent with the assumptions in the data assimilation method. To accomplish this, a discriminator $f_{dis}$ is combined with the autoencoder to force the distribution of the latent variables $\Bxi$ to follow the (target) standard normal distribution $\Bxi \sim N(\textbf{0}, I)$. In the data assimilation process, the trained decoder is applied with ESMDA to generate posterior samples.

In the AAE, shown in Fig.~\ref{fig:aae}(b), the high-dimensional data vector $\Bd^{\text{sim}}$ is input to the encoder to generate the latent vector $\Bxi$. The encoding distribution is denoted $q(\Bxi|\Bd^{\text{sim}})$. The latent vector $\Bxi$ is fed into the decoder to provide the reconstructed data vector $\Bdhat^{\text{sim}}$. The decoding distribution is $q(\Bd^{\text{sim}} | \Bxi)$. The aggregated distribution $q(\Bxi)$ of latent variables is given by 
\begin{equation}\label{eq:enc_q_xi_prob}
    q(\Bxi) = \int_{\Bd^{\text{sim}}} q(\Bxi|\Bd^{\text{sim}}) p(\Bd^{\text{sim}}) d\Bd^{\text{sim}}.
\end{equation}
The target (prior) distribution of the latent variables is denoted as $p(\Bxi)$. The discriminator acts in conjunction with the encoder, with the goal of driving the aggregated posterior distribution $q(\Bxi)$ in Eq.~\ref{eq:enc_q_xi_prob} to match the specified prior distribution (standard normal) $p(\Bxi)$. As shown in Fig.~\ref{fig:aae}(b), the discriminator is applied to predict whether the input $\Bxi$ is from the encoding net $q(\Bxi)$ (false sample), or from the standard normal distribution $p(\Bxi)$ (real sample).

Table~\ref{tab:aae} presents the detailed architecture of the AAE and 3D convLSTM. The input of the autoencoder is of dimension $n_x \times n_y \times n_z \times 2 \times n_t$ (recall $n_t$ is the number of time steps in the data vector, not in the full simulation). The two channels in the data vector correspond to the pressure and saturation fields. The encoder comprises six stacked convLSTM layers with different numbers of filters and strides. The encoder outputs the low-dimensional latent variables of dimensions $\frac{n_x}{16} \times \frac{n_y}{16} \times \frac{n_z}{4} \times 8$. The latent variables are fed to both the decoder and the discriminator. The decoder includes five deconvolutional (upsampling) layers and seven convLSTM layers. The decoder can reconstruct or generate new high-dimensional data vectors of dimension $n_x \times n_y \times n_z \times 2 \times n_t$. The discriminator contains two 3D convolutional layers to further reduce the dimensions of the latent variables. Three dense layers are then attached to output the probability of classifying a sample as false or real. 

We performed hyperparameter tuning to determine appropriate architecture and training details, e.g., the number of stacked convLSTM layers, the learning rate schedule, the batch size, etc. The detailed parameter specifications had relatively little effect on model performance over the range evaluated. We found the general architecture to be quite robust in terms of its ability to parameterize the spatio-temporal data associated with our problem. The settings presented in Table~\ref{tab:aae} provided the best results achieved during hyperparameter tuning. 

We now describe the training procedure. Training provides the parameters associated with the encoder ($W_{enc}$), the decoder ($W_{dec}$), and the discriminator ($W_{dis}$). The data vector ${\bf d}^\text{sim}$ includes both pressure and saturation data. Normalization is, in general, necessary prior to training. Min-max normalization is applied for pressure, with the maximum and minimum values taken to be those observed over the entire field at all time steps. Saturation can be treated directly as its range is already from 0 to 1. The AAE is trained in two phases -- the reconstruction phase and the regularization phase -- in each training batch, as in \cite{makhzani2015adversarial}. In the reconstruction phase, the encoder and decoder are trained jointly to minimize the reconstruction loss, given by
\begin{equation}
    L_{rec} = \frac{1}{n_{smp}}\sum_{i=1}^{n_{smp}}||\Bd^\text{sim}_i - \Bdhat^\text{sim}_i||_2^2,
\end{equation}
where $n_{smp}$ is the total number of the training samples, and the $\Bd^\text{sim}_i$ and $\Bdhat^\text{sim}_i$ vectors are normalized.  Additional weight can be added for data in specific locations (such as near wells). Only network parameters in the encoder and decoder are updated during this training phase. 

In the regularization phase, the discriminator $f_{dis}$ is updated first to distinguish real samples $\Bxi$ (from the standard normal distribution) from false samples $f_{enc}(\Bd^\text{sim})$ generated by the encoder. The discriminator loss is given by  
\begin{equation}
    L_{dis} = -\frac{1}{n_{smp}}\sum_{i=1}^{n_{smp}}\{\log[f_{dis}(\Bxi_i)] + \log[1-f_{dis}\left(f_{enc}(\Bd^\text{sim}_i)\right)]\}.
\end{equation}
The probability of predicting real sample $\Bxi$ and false sample $f_{enc}(\Bd^\text{sim}_i)$ is maximized to improve discriminator performance.

The encoder is then updated to ``fool" the discriminator by increasing the probability of predicting false samples $f_{enc}(\Bd^\text{sim})$ as real samples. The loss in this training step is defined as 
\begin{equation}
    L_{enc} = -\frac{1}{n_{smp}}\sum_{i=1}^{n_{smp}}\log[f_{dis}\left(f_{enc}(\Bd^\text{sim}_i)\right)].
\end{equation}
The regularization phase, which entails training the discriminator and encoder sequentially, allows the AAE to approximate the specified prior distribution of the latent variables. With this approach, the distribution  from the encoder ($q(\Bxi)$ in Eq.~\ref{eq:enc_q_xi_prob}) will gradually approach the specified standard normal distribution $p(\Bxi)$. The decoder is then able to generate new data realizations that follow the features of the prior distributions of the data variables. The minimization procedure is conducted with the adaptive moment estimation (ADAM) optimizer~\cite{kingma2014adam}. 

\begin{table}[!ht]
\footnotesize
\renewcommand{\arraystretch}{1.2} 
\centering
\caption{Architecture of adversarial autoencoder with 3D convLSTM.} 
\begin{tabular}{l | l| l}
\hline
     Network & Layer & Output  \\
     \hline
     \multirow{7}{*}{Encoder ($W_{enc}$)} & Input & $(n_x, n_y, n_z, 2, n_t)$\\
     & convLSTM3D, 16 filters of size $3\times 3 \times 3$, stride 2 & $(\frac{n_x}{2}, \frac{n_y}{2}, \frac{n_z}{2}, 16, n_t)$\\
     & convLSTM3D, 32 filters of size $3\times 3 \times 3$, stride 2 & $(\frac{n_x}{4}, \frac{n_y}{4}, \frac{n_z}{4}, 32, n_t)$\\
     & convLSTM3D, 32 filters of size $3\times 3 \times 3$, stride $2\times 2 \times 1$ & $(\frac{n_x}{8}, \frac{n_y}{8}, \frac{n_z}{4}, 32, n_t)$\\
     & convLSTM3D, 64 filters of size $3\times 3 \times 3$, stride $2\times 2 \times 1$ & $(\frac{n_x}{16}, \frac{n_y}{16}, \frac{n_z}{4}, 64, n_t)$\\
     & convLSTM3D, 16 filters of size $3\times 3 \times 3$, stride 1 & $(\frac{n_x}{16}, \frac{n_y}{16}, \frac{n_z}{4}, 16, n_t)$\\
     & convLSTM3D, 8 filters of size $3\times 3 \times 3$, stride 1 & $(\frac{n_x}{16}, \frac{n_y}{16}, \frac{n_z}{4}, 8)$\\ 
     \hline
     \multirow{14}{*}{Decoder ($W_{dec}$)} 
     & Input & $(\frac{n_x}{16}, \frac{n_y}{16}, \frac{n_z}{4}, 8)$\\ 
     & Repeat vector and reshape & $(\frac{n_x}{16}, \frac{n_y}{16}, \frac{n_z}{4}, 8, n_t)$\\
     & convLSTM3D, 32 filters of size $3\times 3 \times 3$, stride 1 & $(\frac{n_x}{16}, \frac{n_y}{16}, \frac{n_z}{4}, 32, n_t)$\\ 
     & deconv, 32 filters of size $3\times 3 \times 3$, stride 2 & $(\frac{n_x}{8}, \frac{n_y}{8}, \frac{n_z}{2}, 32, n_t)$\\
     & convLSTM3D, 64 filters of size $3\times 3 \times 3$, stride 1 & $(\frac{n_x}{8}, \frac{n_y}{8}, \frac{n_z}{2}, 64, n_t)$ \\ 
     & deconv, 64 filters of size $3\times 3 \times 3$, stride 2 &  $(\frac{n_x}{4}, \frac{n_y}{4}, n_z, 64, n_t)$\\
     & convLSTM3D, 64 filters of size $3\times 3 \times 3$, stride 1 & $(\frac{n_x}{8}, \frac{n_y}{8}, \frac{n_z}{2}, 64, n_t)$ \\ 
     & deconv, 64 filters of size $3\times 3 \times 3$, stride 2 &  $(\frac{n_x}{4}, \frac{n_y}{4}, n_z, 64, n_t)$\\
     & convLSTM3D, 32 filters of size $3\times 3 \times 3$, stride 1 & $(\frac{n_x}{4}, \frac{n_y}{4}, n_z, 32, n_t)$ \\ 
     & deconv, 32 filters of size $3\times 3 \times 3$, stride  $2\times 2 \times 1$ &  $(\frac{n_x}{2}, \frac{n_y}{2}, n_z, 32, n_t)$ \\
     & convLSTM3D, 16 filters of size $3\times 3 \times 3$, stride 1 & $(\frac{n_x}{4}, \frac{n_y}{4}, n_z, 16, n_t)$ \\ 
     & deconv, 16 filters of size $3\times 3 \times 3$, stride  $2\times 2 \times 1$ &  $(n_x, n_y, n_z, 16, n_t)$ \\
     & convLSTM3D, 16 filters of size $3\times 3 \times 3$, stride 1 & $(n_x, n_y, n_z, 16, n_t)$ \\ 
     & convLSTM3D, 2 filters of size $3\times 3 \times 3$, stride 1 & $(n_x, n_y, n_z, 2, n_t)$ \\ 
     \hline
     \multirow{7}{*}{Discriminator ($W_{dis}$)} & Input & $(\frac{n_x}{16}, \frac{n_y}{16}, \frac{n_z}{4}, 8)$ \\
     & conv3D, 32 filters of size $3\times 3 \times 3$, stride 1 & $(\frac{n_x}{16}, \frac{n_y}{16}, \frac{n_z}{4}, 32)$\\
     & conv3D, 16 filters of size $3\times 3 \times 3$, stride 1 & $(\frac{n_x}{16}, \frac{n_y}{16}, \frac{n_z}{4}, 16)$\\
     & Flatten & $(\frac{n_x}{16} \times \frac{n_y}{16} \times \frac{n_z}{4} \times 16, 1)$\\
     & Dense & (128, 1) \\
     & Dense & (64, 1) \\
     & Dense (activation function: $\text{sigmoid}$) & (1, 1) \\
     \hline
\end{tabular}
\label{tab:aae}
\end{table}

\subsection{DSI with data parameterization}
Posterior sampling is combined, within the DSI framework, with the parameterization described above. This enables us to generate posterior forecasts of the pressure and saturation fields at a set of time steps. \citet{jiang2021data} combined deep-learning parameterization methods with ESMDA for well-rate predictions (not to predict global pressure or saturation fields) in DSI. With ESMDA, the observations are assimilated multiple times and the posterior distributions are updated based on the covariance between (observed) historical data and the variables to be calibrated. ESMDA enables efficient posterior sampling, and it was shown to provide posterior DSI statistics comparable to those from the time-consuming optimization-based randomized maximum likelihood (RML) method in \cite{jiang2021data}.  

In the DSI framework with parameterization, the latent variables $\Bxi$ are updated. The mapping of the latent variables to data variables (performed in the decoder), $\Bd^\text{sim} \approx \Bdhat^\text{sim} = f_{dec}(\Bxi)$, is applied to generate the simulation data required in ESMDA. The ESMDA update for the latent variables is given by \cite{jiang2021data} 
\begin{equation}\label{eq:esmda_ksi}
\Bxi_i^{k+1} = \Bxi_i^{k} + C_{\xi, \Bdhat^\text{sim}_{\text{hm}}}^k(C_{\Bdhat^\text{sim}_{\text{hm}}}^k + \alpha_k\CD)^{-1}(\Bdobs^\text{tot}+\sqrt{\alpha_k}\mathbf{e}_i^k - (\Bdhat_\text{hm}^\text{sim})_i^k),
\end{equation}
for $k = 1, \ldots, \Na$ and $i = 1, \ldots, \Nr$, where $N_a$ is the number of assimilation steps. Here $\alpha_k$ are the inflation coefficients, $C_{\Bdhat^\text{sim}_{\text{hm}}}$ is the covariance matrix of historical data variables $\Bdhat_\text{hm}^\text{sim}$, and $C_{\xi, \Bdhat^\text{sim}_{\text{hm}}}$ is the cross-covariance matrix between $\Bxi$ and $\Bdhat_\text{hm}^\text{sim}$. The historical period data $(\Bdhat_\text{hm}^\text{sim})_i^k$ are generated via $(\Bdhat_\text{hm}^\text{sim})_i^k = E f_{dec}(\Bxi_i^k)$ at every iteration $k$. The covariance matrices $C_{\xi, \Bdhat^\text{sim}_{\text{hm}}}^k$ and $C_{\Bdhat^\text{sim}_{\text{hm}}}^k$ are then updated with the new ensemble of $(\Bdhat_{\text{hm}}^\text{sim})^k$ and $\Bxi^{k}$. In this work, we set $\Na=4$ and $\alpha_k = 4$ for $k = 1, \ldots, 4$. This specification satisfies $\sum_{k=1}^{\Na}\alpha_k^{-1} = 1$, as given in \cite{emerick2013ensemble}. After $\Na$ ESMDA iterations are completed, posterior predictions are generated through application of $(\Bdhat_{\text{post}})_i = f_{dec} \left( (\Bxi_{\text{post}})_i \right)$, $i = 1, \ldots, \Nr$, where $\Bxi_{\text{post}}$ denotes the final posterior latent variables.
\section{Reproduction of prior results and statistics}\label{sec:prior}
In this section, we first describe the aquifer model and simulation setup. We then show reconstructed pressure and saturation fields using the new AAE with convLSTM parameterization. Prior statistics for pressure and saturation for new realizations, generated with the AAE from Gaussian-distributed latent variables, are then presented.  

\subsection{Aquifer model}
\begin{figure}[!htb]
\centering
\begin{minipage}{.4\linewidth}\centering
\includegraphics[trim = 0 0 100 0, clip, width=\linewidth]{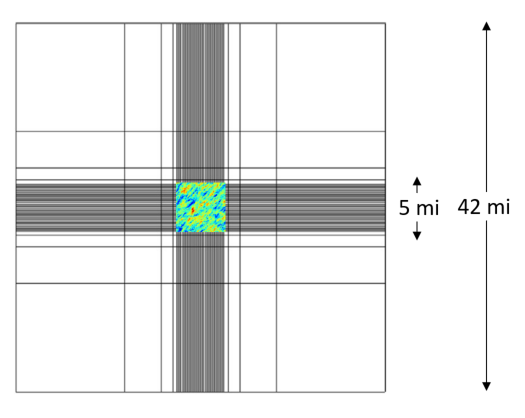}
\subcaption{Full domain including surrounding region, with storage aquifer shown in the center}
\end{minipage}
\hfill
\begin{minipage}{.45\linewidth}\centering
\includegraphics[trim = 10 10 10 10, clip, width=\linewidth]{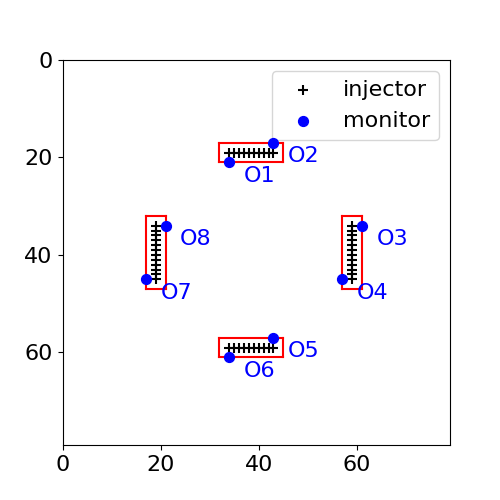}
\subcaption{Locations of four injection wells and eight observation wells in the storage aquifer}
\end{minipage}
\caption{Model domains for flow simulations.} \label{fig:geomodel}
\end{figure}

We consider systems of a size suitable for industrial-scale carbon storage operations. The full model, shown in Fig.~\ref{fig:geomodel}a, is of size 67~km $\times$ 67~km $\times$ 122~m. The domain includes a storage aquifer, in the center of the model, and a surrounding region to capture large-scale pressure effects. The storage aquifer is of size 8.5~km $\times$ 8.5~km $\times$ 122~m and is represented on a uniform grid containing 80 $\times$ 80 $\times$ 20 cells. The grid-block size increases with distance away from the storage aquifer, as shown in Fig.~\ref{fig:geomodel}a. A high degree of resolution is not required in the surrounding region, and this approach provides computational savings. The full model contains 88 $\times$ 88 $\times$ 20 cells (total of 154,880 cells).

The geological properties of the storage aquifer represent generalizations of the models presented in \citet{okwen2022storage} and \citet{crain2023integrated}. These models are for a carbon storage project planned for the Mt.~Simon formation in central Illinois, USA. The geomodels in \citet{crain2023integrated} contained six geological layers with different statistical properties. The uppermost layer was a shale layer, which had essentially no impact on flow. We therefore do not include this layer in our models.

In this work, we treat a much wider range of geomodel realizations than was considered by \citet{crain2023integrated}. Specifically, we construct realizations corresponding to a broad distribution of geological scenarios, each characterized by a set of scenario parameters (also referred to as metaparameters). In \cite{crain2023integrated}, by contrast, a single scenario (i.e., one set of metaparameters) was considered. For any particular set of metaparameters, an infinite number of multi-Gaussian-distributed realizations can be generated. Realizations are constructed here using sequential Gaussian simulation within the geostatistical toolbox SGeMS~\cite{remy2009applied}. The metaparameters, shown in Table~\ref{tab:geomodel_para}, include, for each geological layer, the mean and standard deviation of log-permeability and porosity ($\mu_{\log k}$, $\sigma_{\log k}$, $\mu_{\phi}$, $\sigma_{\phi}$), correlation lengths in three directions ($l_x$, $l_y$, $l_z$), variogram orientation (azimuth and dip), and permeability anisotropy ratio ($k_z/k_x)$. The number of simulation layers used to represent each geological layer ($n_z$) is also shown in the table.

In our modeling, the metaparameters are sampled from uniform distributions over the ranges given in Table~\ref{tab:geomodel_para}. For a particular geological layer, the sampled value of $k_z/k_x$ is applied to all grid blocks in that layer. Exponential variogram models are used, and in all cases we specify $k_x=k_y$. In the region outside the storage aquifer, the permeability and porosity are set to be constant at 10~md and 0.2. The ranges in Table~\ref{tab:geomodel_para} include the parameter values used in \cite{crain2023integrated}. The consideration of distributions for the metaparameters adds challenges for both the AAE parameterization and for history matching. 

\begin{table}[!ht] 
\begin{center}
\footnotesize
\caption{Parameter ranges for each geological layer. All parameters are uniformly distributed.}\label{tab:geomodel_para}
\renewcommand{\arraystretch}{1.2} 
\begin{tabular}{cccccc} 
\hline
\text{Parameters} & \text{Layer~1} & \text{Layer~2}  & \text{Layer~3}  & \text{Layer~4}  & \text{Layer~5} \\ 
\hline
$n_z$ & 5 & 4 & 5 & 3 & 3\\
$\mu_{\log k}$  &[2, 3] & [1.5, 2.5] & [2, 3] & [3, 5] & [2, 3]\\ 
$\sigma_{\log k}$ & [1, 2] & [1.5, 2.5] & [1, 2] & [1.5, 2.5] & [1.5, 2.5]  \\ 
$\mu_{\phi}$  &[0.175, 0.225] & [0.15, 0.2] & [0.175, 0.225] & [0.25, 0.3] & [0.175, 0.225]\\ 
$\sigma_{\phi}$ & [0.025, 0.03] & [0.02, 0.025] & [0.025, 0.03] & [0.035, 0.04] & [0.025, 0.03] \\ 
$l_x$ & [10, 30] & [10, 20] & [20, 40] & [10, 30] & [10, 30] \\
$l_y$ & [10, 30] & [10, 20] & [20, 40] & [10, 30] & [10, 30] \\
$l_z$ &  [3, 5] & [2, 4] & [3, 5] & [1, 3] & [10, 30]\\
Azimuth & [30, 45] & [30, 45] & [30, 45] & [30, 45] & [30, 45]\\
Dip & [30, 45] & [30, 45] & [30, 45] & [30, 45] & [30, 45]\\
$k_z / k_x$  & [0.05, 0.2] & [0.05, 0.2] & [0.05, 0.2] & [0.05, 0.2] & [0.05, 0.2]\\
\hline
\end{tabular}
\end{center}
\end{table}

\begin{figure}[!htb]
\centering
\begin{minipage}{.327\linewidth}\centering
\includegraphics[trim = 0 0 0 0, clip, width=\linewidth]{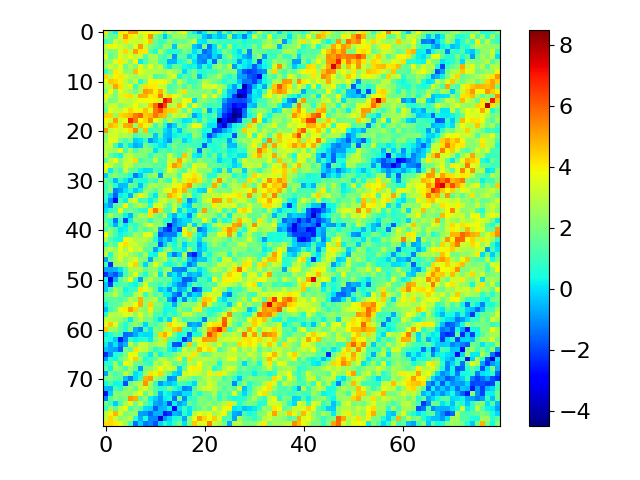}
\subcaption{Prior realization~1}
\end{minipage}
\begin{minipage}{.327\linewidth}\centering
\includegraphics[trim = 0 0 0 0, clip, width=\linewidth]{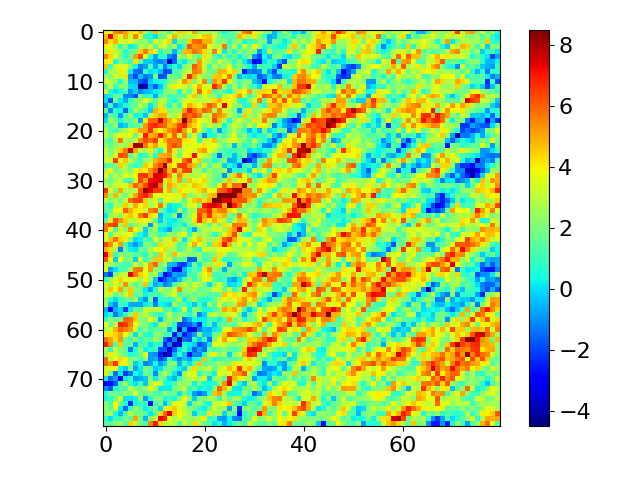}
\subcaption{Prior realization~2}
\end{minipage}
\begin{minipage}{.327\linewidth}\centering
\includegraphics[trim = 0 0 0 0, clip, width=\linewidth]{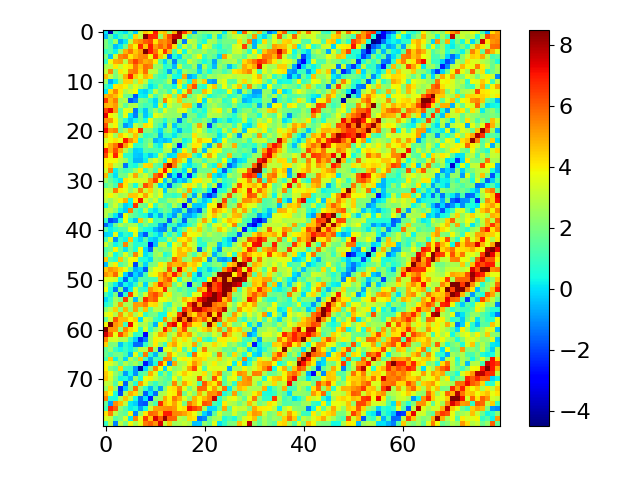}
\subcaption{Prior realization~3}
\end{minipage}
\caption{Log-permeability realizations of three prior models for simulation layer~18. Realizations in (a) and (b) correspond to true models~1 and~2 used for history matching.} \label{fig:perm_prior}
\end{figure}

Three random realizations of log-permeability for simulation layer~18 are presented in Fig.~\ref{fig:perm_prior}. The model in Fig.~\ref{fig:perm_prior}a, taken to be `true' model~1, is used in many of the results. The model in Fig.~\ref{fig:perm_prior}b is true model~2, which is used later for history matching. Clear differences between these three realizations in average permeability and correlation structure are evident. 

In the flow simulations, CO$_2$ is injected through four horizontal wells located in layer~18 (top layer of geological layer~5). Each well injects 1~Mt/year, so a total of 4~Mt/year is injected. Injection proceeds for 20~years. The injectors are shown as the black lines in Fig.~\ref{fig:geomodel}b. Eight vertical monitoring/observation wells, denoted O1--O8, penetrate the entire aquifer. The red rectangular boxes depict local grid refinement (LGR) regions. These will be used later to estimate model resolution error, which enters into the history matching procedure. The monitoring wells in this study are placed in locations generally consistent with those in \cite{sun2019data}. The monitoring well locations could be optimized using the procedure in \cite{sun2019data}, though this entails the specification of a particular objective function to be minimized. Because we have many different quantities of interest in this work, the appropriate objective function (or set of weighted objective functions) would require some investigation.

\begin{figure}[!htb]
\centering
\begin{minipage}{.45\linewidth}\centering
\includegraphics[trim = 0 0 0 0, clip, width=\linewidth]{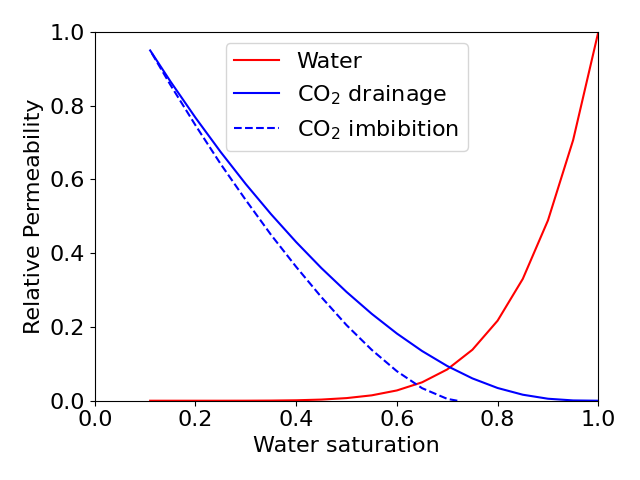}
\subcaption{Relative permeability curves}
\end{minipage}
\begin{minipage}{.45\linewidth}\centering
\includegraphics[trim = 0 0 0 0, clip, width=\linewidth]{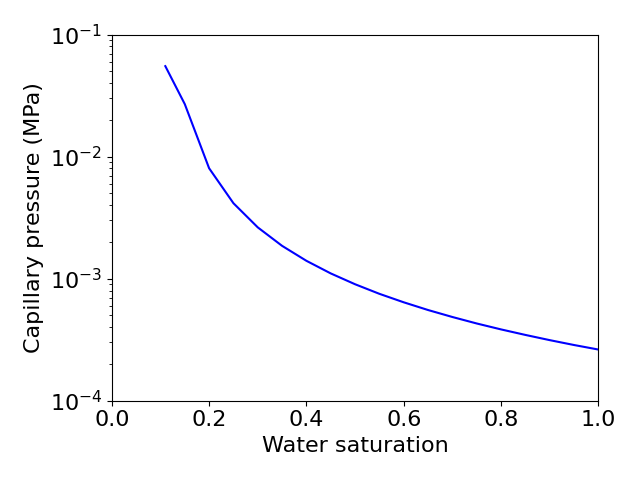}
\subcaption{Capillary pressure curve for $k=10$~md, $\phi=0.2$} 
\end{minipage}
\caption{Two-phase flow functions (after \cite{cameron2012optimization}).} \label{fig:flow-properties}
\end{figure}

The flow simulations are performed using the ECLIPSE simulator with the CO2STORE option~\cite{Schlumberger}. Two fluid phases (water and gas) and three components (water, supercritical CO$_2$ and salt) are considered. The initial pressure of the storage aquifer for the top layer is 15.5~MPa and the temperature is 55~$^\circ$C. No-flow boundary conditions are applied at the boundaries of the full system. The gas-water relative permeability curves, which include hysteresis, are shown in Fig.~\ref{fig:flow-properties}a. The Brooks-Corey model~\cite{saadatpoor2010new} with entry pressure ($p_e$) of 0.36 and exponent ($\lambda$) of 0.67 is used to generate the capillary pressure curves. The Leverett $J$-function is applied to calculate the capillary pressure for each grid block based on the porosity and permeability. Figure~\ref{fig:flow-properties}b presents the capillary pressure curve at porosity 0.2 and permeability 10~md. 

\subsection{Reconstruction of prior results}

A total of 2000 prior realizations, drawn from multiple geological scenarios, are simulated. The pressure and saturation are collected at $n_t = 6$ time steps, corresponding to years~1, 3, 5, 10, 15, 20 after the start of injection. The AAE model is trained with 1500 realizations. The other 500 realizations comprise the test set used to evaluate the performance of the parameterization method and for history matching. The AAE parameterization represents the high-dimensional dynamic pressure and saturation fields with 1000 latent variables. The AAE model includes 2,976,665 parameters in total. The overall training time is 11 hours for 200 epochs using a single Nvidia Tesla A100 GPU.

\begin{figure}[!htb]
\centering
\begin{minipage}{.32\linewidth}\centering
\includegraphics[trim = 0 0 0 0, clip, width=\linewidth]{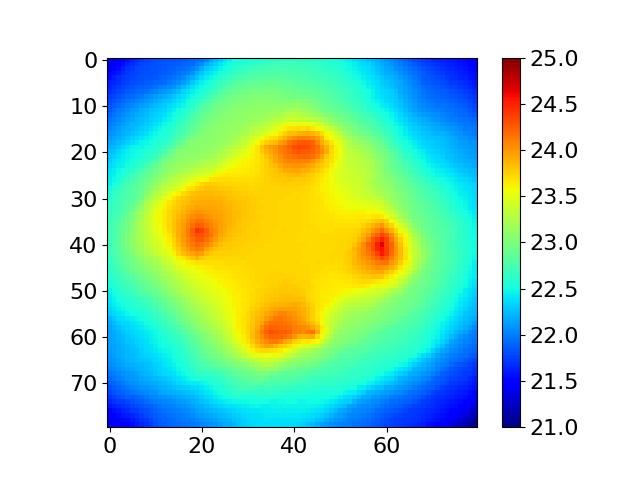}
\subcaption{Realization~1 (sim)}
\end{minipage}
\begin{minipage}{.32\linewidth}\centering
\includegraphics[trim = 0 0 0 0, clip, width=\linewidth]{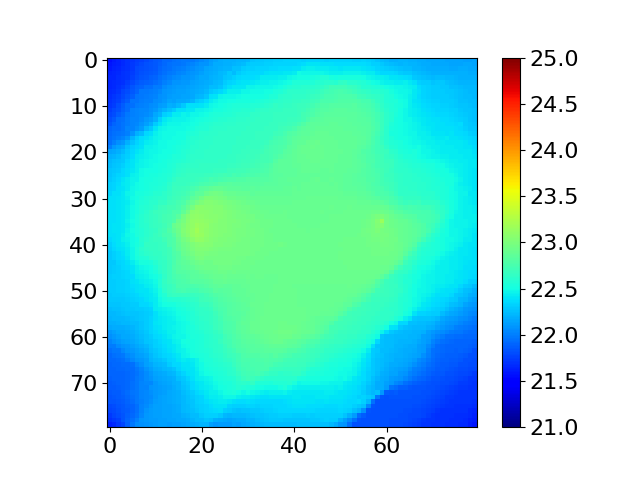}
\subcaption{Realization~2 (sim)}
\end{minipage}
\begin{minipage}{.32\linewidth}\centering
\includegraphics[trim = 0 0 0 0, clip, width=\linewidth]{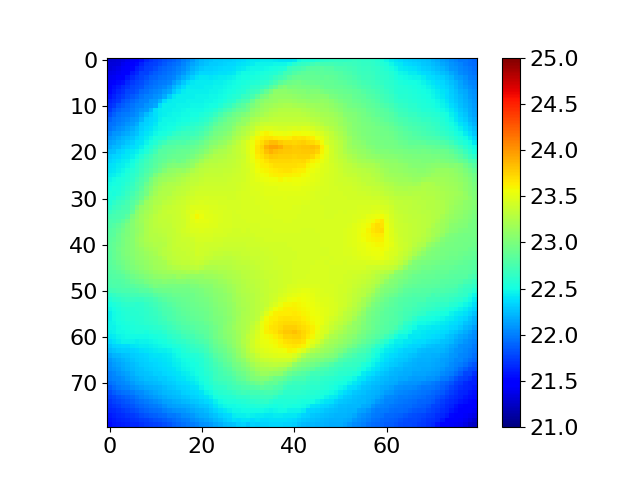}
\subcaption{Realization~3 (sim)}
\end{minipage}
\begin{minipage}{.32\linewidth}\centering
\includegraphics[trim = 0 0 0 0, clip, width=\linewidth]{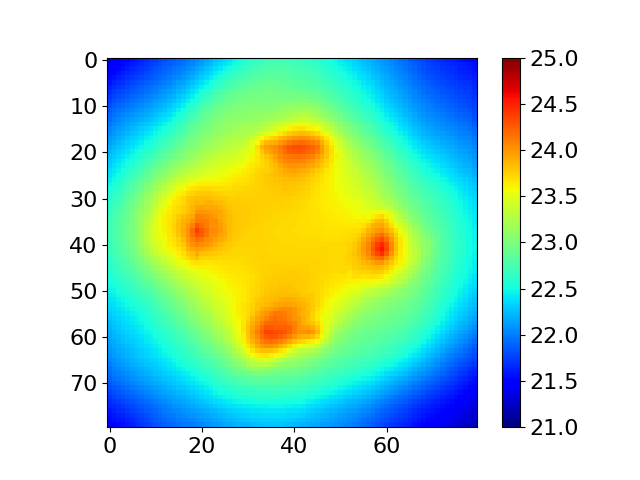}
\subcaption{Realization~1 (AAE)}
\end{minipage}
\begin{minipage}{.32\linewidth}\centering
\includegraphics[trim = 0 0 0 0, clip, width=\linewidth]{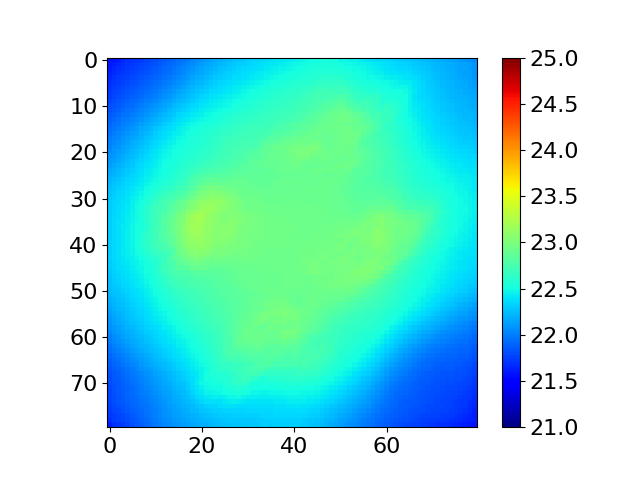}
\subcaption{Realization~2 (AAE)}
\end{minipage}
\begin{minipage}{.32\linewidth}\centering
\includegraphics[trim = 0 0 0 0, clip, width=\linewidth]{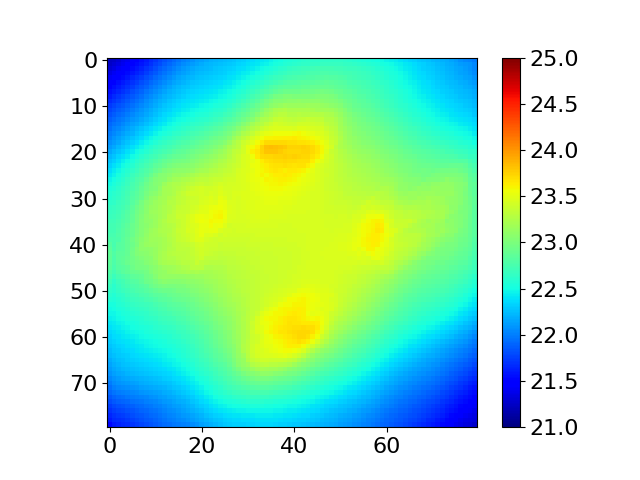}
\subcaption{Realization~3 (AAE)}
\end{minipage}
\caption{Pressure fields in the storage aquifer for flow simulation (upper row) and the AAE model (lower row) for the three geomodels in Fig.~\ref{fig:perm_prior}. Results are for simulation layer~15 at 20~years.} \label{fig:prior_p_map}
\end{figure}

We first evaluate the performance of the new AAE treatment by comparing reconstructed pressure and saturation fields to reference simulation results. The upper row of Fig.~\ref{fig:prior_p_map} displays simulation results for pressure, for simulation layer~15 at 20~years, for the three realizations in Fig.~\ref{fig:perm_prior}. The lower row shows the corresponding reconstructed pressure fields. The reconstructed fields are in close visual agreement with the simulation results, which demonstrates the ability of the AAE to represent pressure. Clear differences are observed between the three realizations, indicative of the variability between the geological scenarios.  

\begin{figure}[!htb]
\centering
\begin{minipage}{.32\linewidth}\centering
\includegraphics[trim = 0 0 0 0, clip, width=\linewidth]{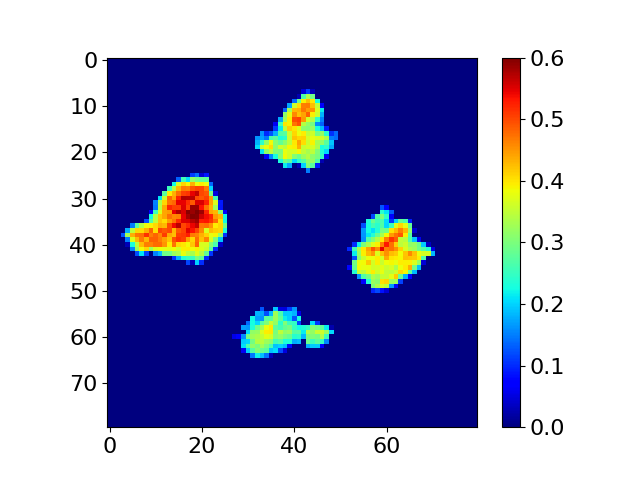}
\subcaption{Realization~1 (sim)}
\end{minipage}
\begin{minipage}{.32\linewidth}\centering
\includegraphics[trim = 0 0 0 0, clip, width=\linewidth]{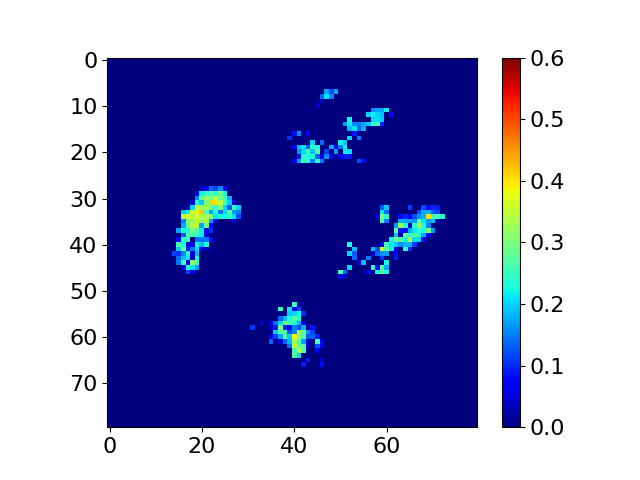}
\subcaption{Realization~2 (sim)}
\end{minipage}
\begin{minipage}{.32\linewidth}\centering
\includegraphics[trim = 0 0 0 0, clip, width=\linewidth]{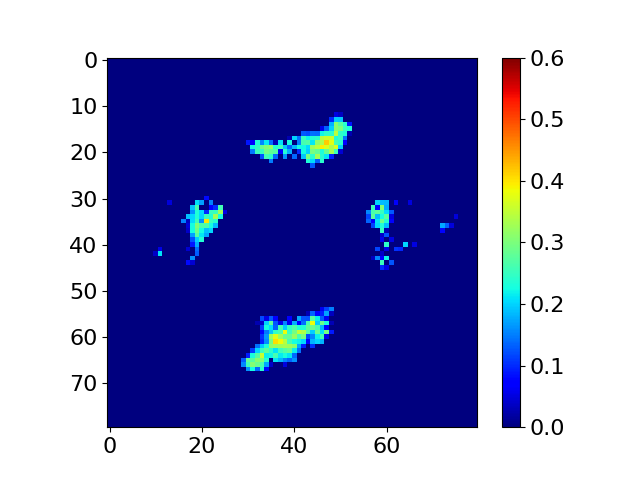}
\subcaption{Realization~3 (sim)}
\end{minipage}
\begin{minipage}{.32\linewidth}\centering
\includegraphics[trim = 0 0 0 0, clip, width=\linewidth]{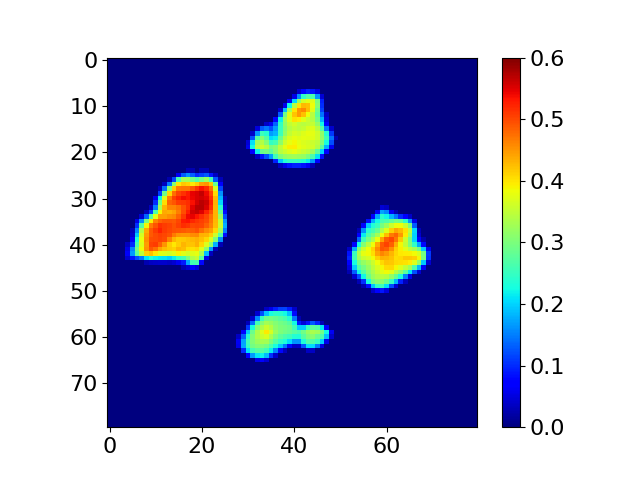}
\subcaption{Realization~1 (AAE)}
\end{minipage}
\begin{minipage}{.32\linewidth}\centering
\includegraphics[trim = 0 0 0 0, clip, width=\linewidth]{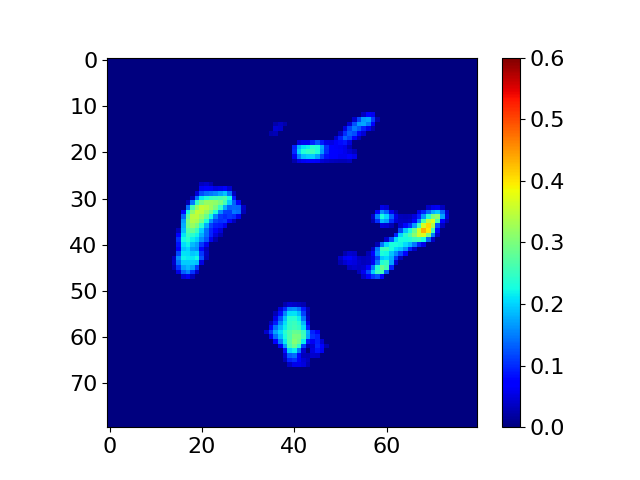}
\subcaption{Realization~2 (AAE)}
\end{minipage}
\begin{minipage}{.32\linewidth}\centering
\includegraphics[trim = 0 0 0 0, clip, width=\linewidth]{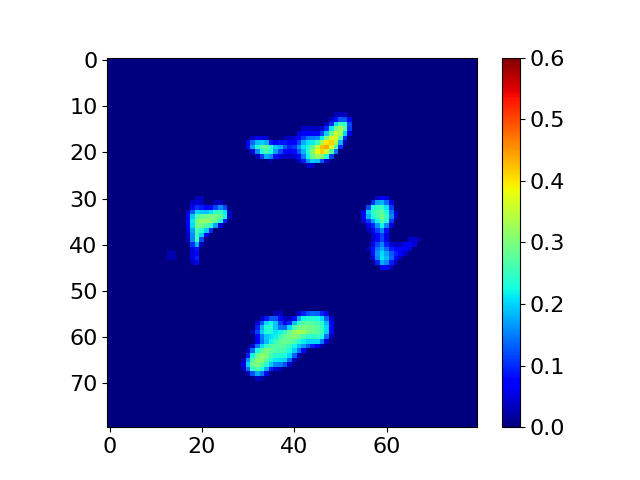}
\subcaption{Realization~3 (AAE)}
\end{minipage}
\caption{CO$_2$ saturation fields in the storage aquifer for flow simulation (upper row) and the AAE model (lower row) for three prior geomodels for simulation layer~10 at 20~years.} \label{fig:prior_s_map_1}
\end{figure}

\begin{figure}[!htb]
\centering
\begin{minipage}{.32\linewidth}\centering
\includegraphics[trim = 0 0 0 0, clip, width=\linewidth]{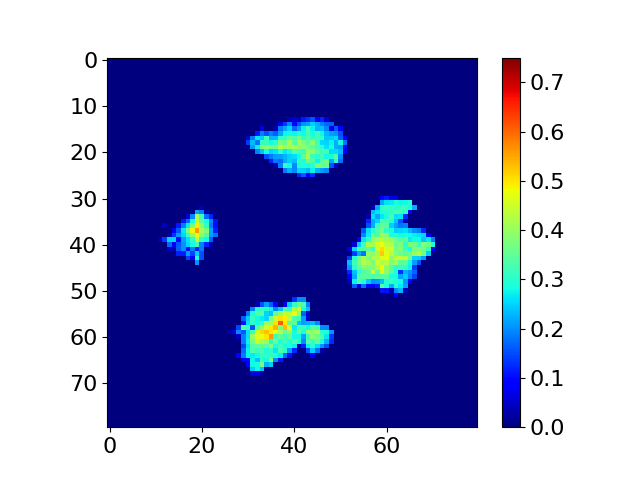}
\subcaption{Realization~1 (sim)}
\end{minipage}
\begin{minipage}{.32\linewidth}\centering
\includegraphics[trim = 0 0 0 0, clip, width=\linewidth]{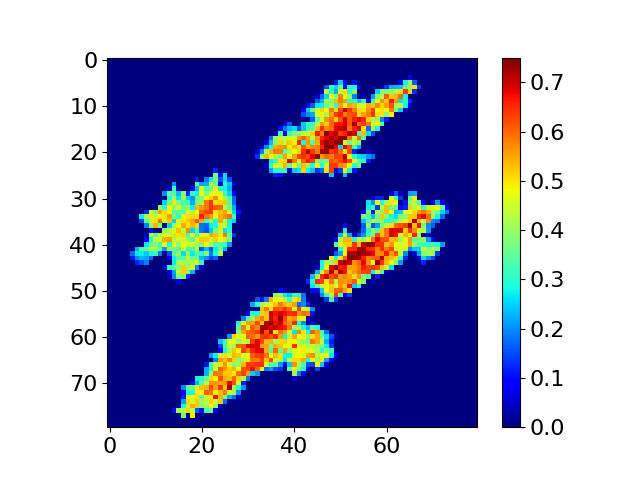}
\subcaption{Realization~2 (sim)}
\end{minipage}
\begin{minipage}{.32\linewidth}\centering
\includegraphics[trim = 0 0 0 0, clip, width=\linewidth]{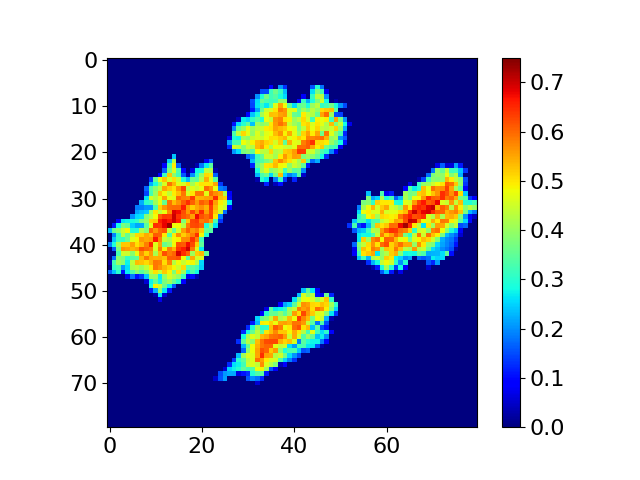}
\subcaption{Realization~3 (sim)}
\end{minipage}
\begin{minipage}{.32\linewidth}\centering
\includegraphics[trim = 0 0 0 0, clip, width=\linewidth]{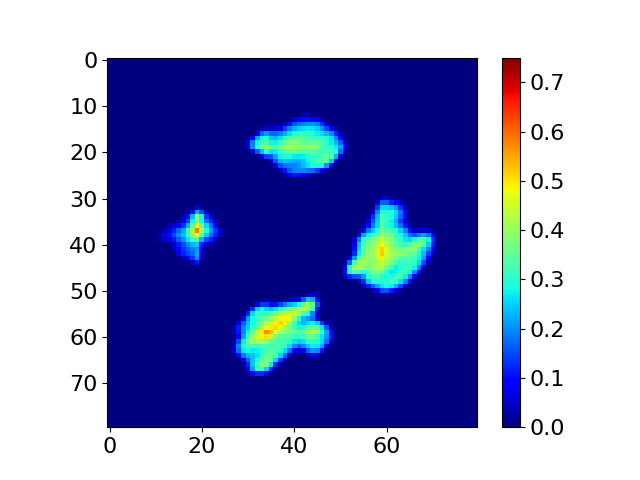}
\subcaption{Realization~1 (AAE)}
\end{minipage}
\begin{minipage}{.32\linewidth}\centering
\includegraphics[trim = 0 0 0 0, clip, width=\linewidth]{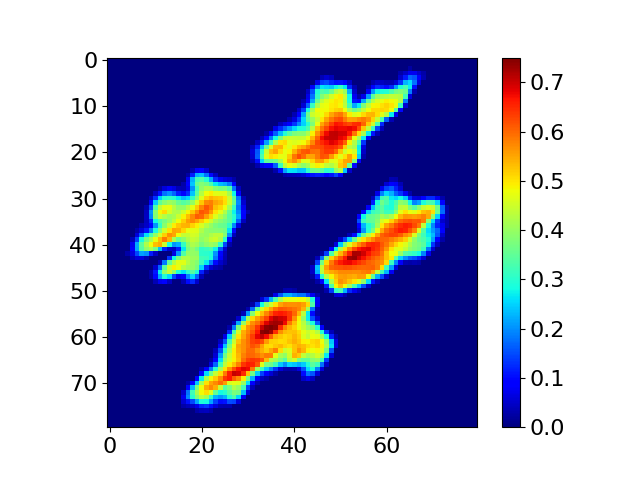}
\subcaption{Realization~2 (AAE)}
\end{minipage}
\begin{minipage}{.32\linewidth}\centering
\includegraphics[trim = 0 0 0 0, clip, width=\linewidth]{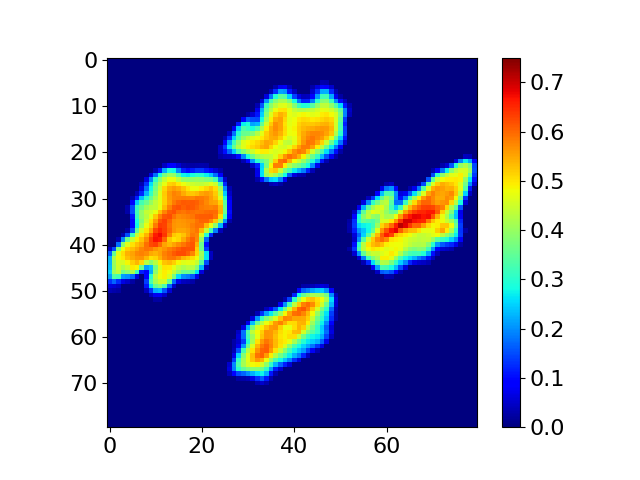}
\subcaption{Realization~3 (AAE)}
\end{minipage}
\caption{CO$_2$ saturation fields in the storage aquifer from flow simulation (upper row) and the AAE model (lower row) for three prior geomodels for simulation layer~15 at 20~years.} \label{fig:prior_s_map_2}
\end{figure}

Figures~\ref{fig:prior_s_map_1} and \ref{fig:prior_s_map_2} present CO$_2$ saturation results for the same three realizations, for simulation layer~10 (the top simulation layer of geological layer~3) and simulation layer~15 (which displays the largest average CO$_2$ saturation for most cases) at 20~years. The CO$_2$ plume generally remains in geological layers~3 and 4 (which correspond to simulation layers~10--17). Significant variability in the saturation fields between the different realizations is observed. The reconstruction results again display general agreement with the reference simulation results, though some discrepancies are evident. In particular, there are slight differences around the saturation fronts as well as in some of the fine-scale variation within the plumes. Nonetheless, the AAE parameterization is clearly able to represent a wide range of plume shapes and saturation distributions. 

\begin{figure}[!hbt]
\centering

\begin{minipage}{.45\linewidth}\centering
\includegraphics[trim = 0 0 0 0, clip, width=\linewidth]{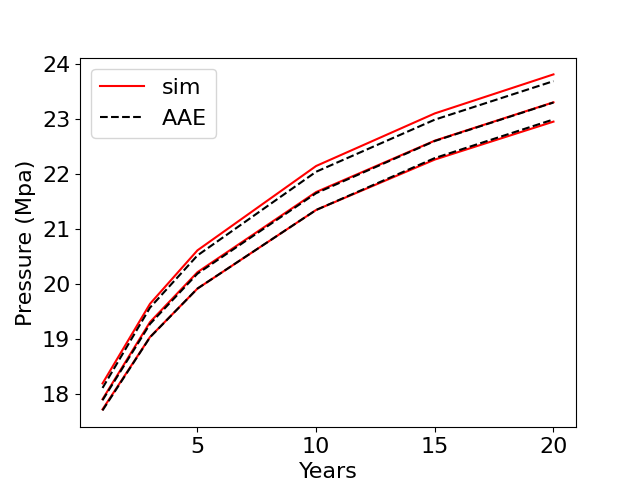}
\subcaption{O7 pressure in layer~15}
\end{minipage}
\begin{minipage}{.45\linewidth}\centering
\includegraphics[trim = 0 0 0 0, clip, width=\linewidth]{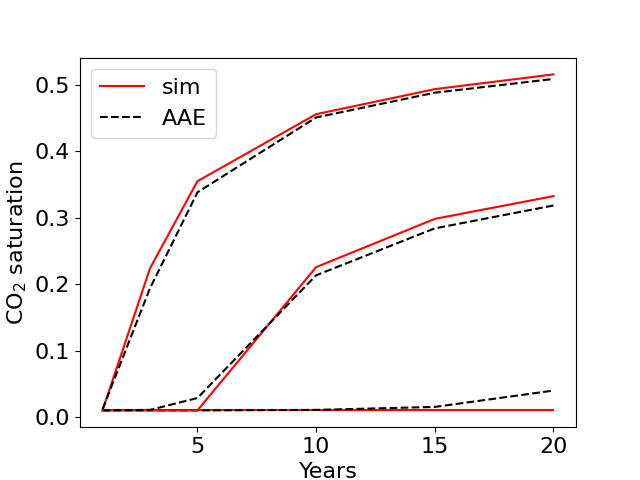}
\subcaption{O7 saturation in layer~15}
\end{minipage}

\begin{minipage}{.45\linewidth}\centering
\includegraphics[trim = 0 0 0 0, clip, width=\linewidth]{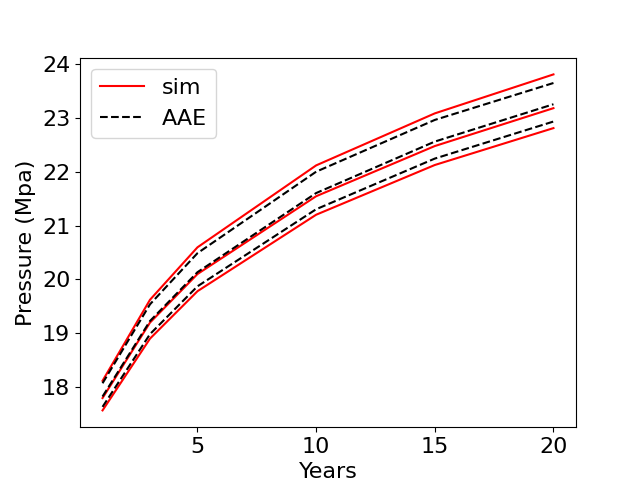}
\subcaption{O8 pressure in layer~10}
\end{minipage}
\begin{minipage}{.45\linewidth}\centering
\includegraphics[trim = 0 0 0 0, clip, width=\linewidth]{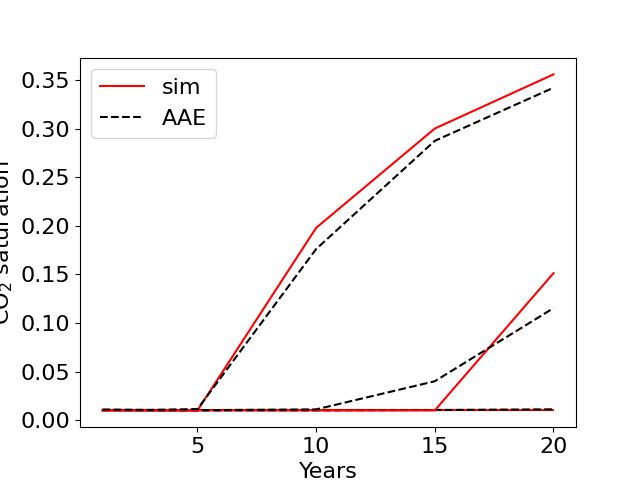}
\subcaption{O8 saturation in layer~10}
\end{minipage}

\caption{Comparison of pressure and saturation statistics derived from flow simulation results (red solid curves) and from new AAE data realizations (black dashed curves). Lower, middle and upper curves are P$_{10}$, P$_{50}$ and P$_{90}$ responses.}\label{fig:prior_flow}
\end{figure}

The AAE parameterization method is constructed to generate data realizations (pressure and saturation fields) from Gaussian-distributed latent variables. To assess performance in this regard, we now compare statistics for new data realizations, generated by inputting 100 samples of standard normal $\Bxi$ to the AAE, to new simulation results. The latter require the construction of new (random) geomodels followed by flow simulation. Figure~\ref{fig:prior_flow} presents the pressure and saturation statistics for observation well~O7 in layer~15 and observation well~O8 in layer~10. In the figures we show the P$_{10}$, P$_{50}$ and P$_{90}$ (10th, 50th and 90th percentile) results over the 20-year injection period. The red lines represent the flow statistics from simulation for the 100 random test samples, while the black dashed lines show the results for pressure and saturation fields generated directly by the AAE. The two sets of results are in reasonable agreement, though some discrepancies are apparent, e.g., the P$_{50}$ saturation in Fig.~\ref{fig:prior_flow}d. The general correspondence in these overall results demonstrates that the AAE is able to generate new data realizations that are statistically consistent with those from (much more expensive) numerical flow simulations. This capability will be very useful for history matching.

\section{Data assimilation results using DSI}\label{sec:post}
In this section, we first describe our estimation of model resolution error, required for history matching, from a local grid refinement assessment. We then present history matching results using the AAE parameterization and ESMDA for posterior sampling in DSI. 

\subsection{Estimation of model error}\label{sec:LGR}
In practice, pressure and saturation data would be measured from instruments in the field, though in this study we use synthetic data that derive from flow simulation performed on a designated true model. The error that appears in history matching formulations includes both measurement error, which results from inaccuracies in the actual instrumentation, and model error. Model error is, in general, challenging to quantify, and many effects can be considered. Following \citet{jiang2021treatment}, model error is here assumed to derive from the mismatch in scale between the observations and flow model resolution. This is an important source of model error though other contributions, such as those from the two-phase flow model or the phase behavior treatment, could also be considered.
 
We apply a grid-refinement study in order to estimate the error associated with model resolution. The specific approach is as follows. LGR representations are introduced in the rectangular regions shown in Fig.~\ref{fig:geomodel}b. These regions extend vertically through the entire model and include the injection wells. Refinement is applied only in the LGR regions, and no additional heterogeneity is introduced as these regions are refined. Three grid levels, corresponding to  $3 \times 3 \times 3$, $5 \times 5 \times 5$, and $7 \times 7 \times 7$ refinements, are considered in this work.
  
\begin{figure}[!htb]
\centering
\begin{minipage}{.24\linewidth}\centering
\includegraphics[trim = 0 80 0 80, clip, width=\linewidth]{ 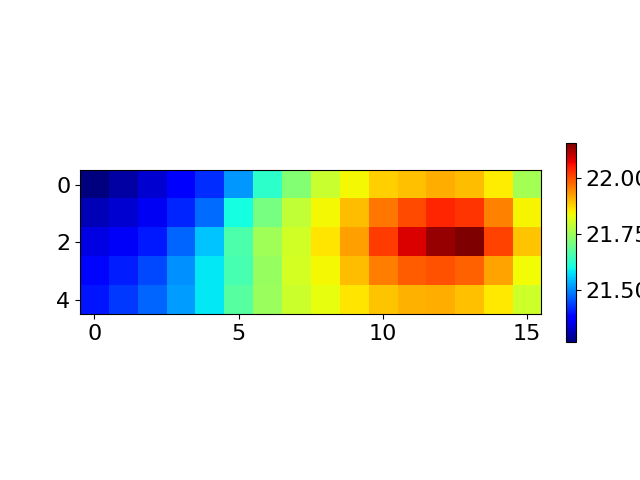}
\subcaption{Low-fidelity pressure}
\end{minipage}
\begin{minipage}{.24\linewidth}\centering
\includegraphics[trim = 0 80 0 80, clip, width=\linewidth]{ 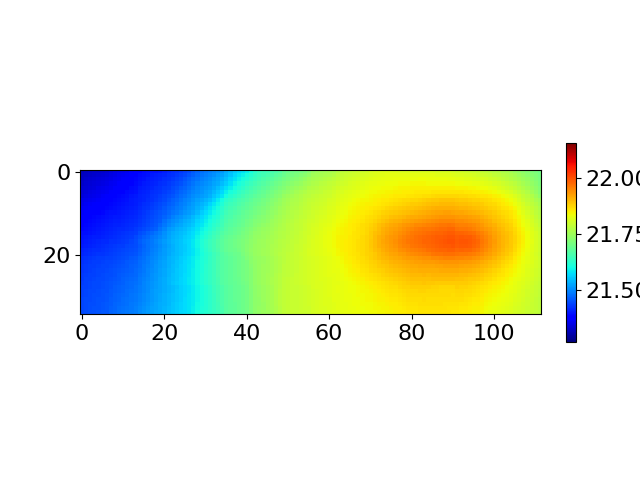}
\subcaption{$3 \times 3 \times 3$}
\end{minipage}
\begin{minipage}{.24\linewidth}\centering
\includegraphics[trim = 0 80 0 80, clip, width=\linewidth]{ 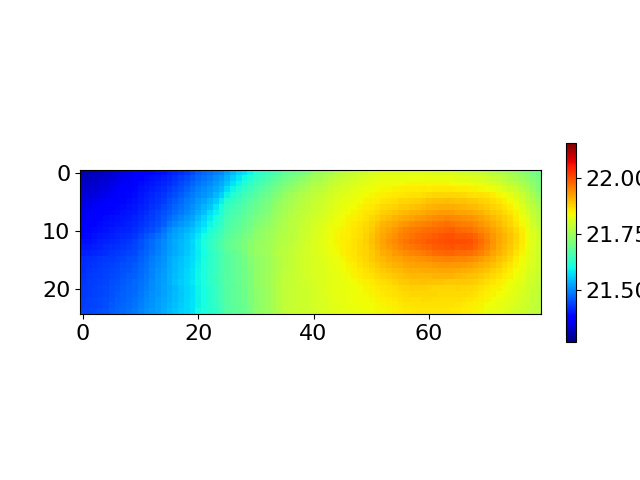}
\subcaption{$5 \times 5 \times 5$}
\end{minipage}
\begin{minipage}{.24\linewidth}\centering
\includegraphics[trim = 0 80 0 80, clip, width=\linewidth]{ pmap_777_1_0.png}
\subcaption{$7 \times 7 \times 7$}
\end{minipage}
\begin{minipage}{.24\linewidth}\centering
\includegraphics[trim = 0 80 0 80, clip, width=\linewidth]{ 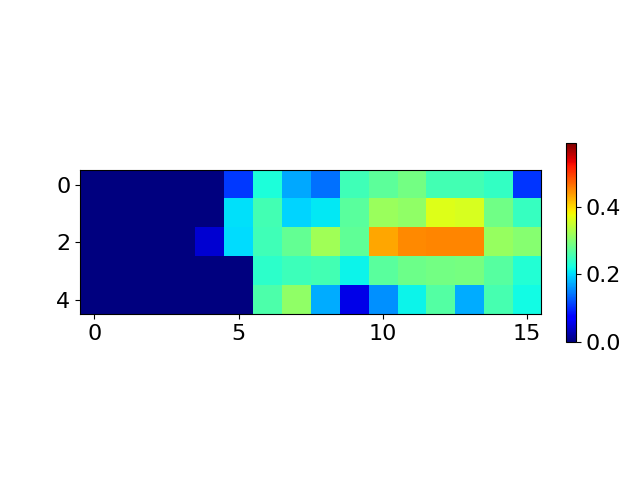}
\subcaption{Low-fidelity saturation}
\end{minipage}
\begin{minipage}{.24\linewidth}\centering
\includegraphics[trim = 0 80 0 80, clip, width=\linewidth]{ 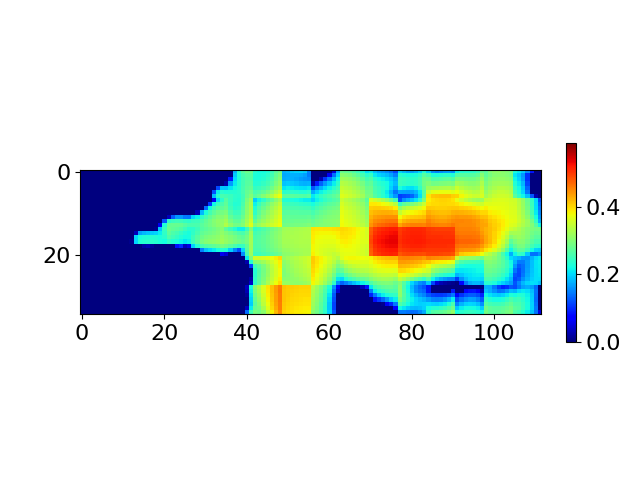}
\subcaption{$3 \times 3 \times 3$}
\end{minipage}
\begin{minipage}{.24\linewidth}\centering
\includegraphics[trim = 0 80 0 80, clip, width=\linewidth]{ 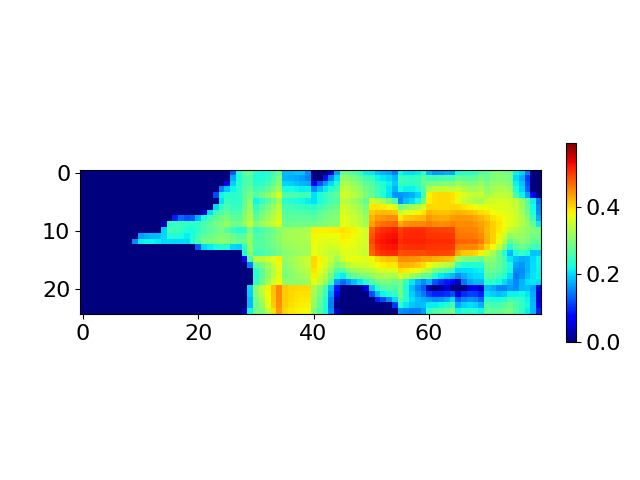}
\subcaption{$5 \times 5 \times 5$}
\end{minipage}
\begin{minipage}{.24\linewidth}\centering
\includegraphics[trim = 0 80 0 80, clip, width=\linewidth]{ smap_777_1_0.png}
\subcaption{$7 \times 7 \times 7$}
\end{minipage}
\caption{Pressure and saturation fields for low-fidelity flow simulations and with local grid refinement at three different refinement levels. Results shown for a random realization for a portion of layer~15 at 10~years.} \label{fig:lgr_fields}
\end{figure}

Figure~\ref{fig:lgr_fields} shows pressure (upper row) and saturation (lower row) results for three refinement levels for a region that includes observation wells~O7 and O8. These results are for a random realization, for a portion of layer~15 at a time of 10~years. The left-most images (Fig.~\ref{fig:lgr_fields}a and e) present the original (lowest fidelity, meaning no LGR) simulation results. Finer-scale effects appear with increasing resolution, particularly in the case of saturation.

The pressure and saturation data used in history matching are at the low-fidelity grid resolution. To quantify the difference between these results and the refined solutions, we now compute the bias and variance of the LGR results relative to the low-fidelity results. Results are reported in Table~\ref{tab:lgr}. The bias for a particular low-fidelity block is computed as the difference between the average value of the LGR results (over the region corresponding to the low-fidelity block) and the low-fidelity simulation result. The biases presented in Table~\ref{tab:lgr} are averages over all such regions. Standard deviations of the LGR results are also computed for each LGR region corresponding to a low-fidelity block. The values in Table~\ref{tab:lgr} are again averages over all such regions. We see that the detailed bias and variance change with grid resolution, but the general magnitude is consistent over the three grid levels. 

The representation of bias in history matching is somewhat complicated since this quantity does not appear directly in the ESMDA equations (see, e.g., \cite{jiang2021treatment} for a detailed treatment within the DSI framework). A treatment for bias does not appear to be essential here, however, since the biases in Table~\ref{tab:lgr} are relatively small. Error standard deviation (or error covariance) does appear directly in the ESMDA equations. For the model error component, we will apply the $7 \times 7 \times 7$ refinement values in Table~\ref{tab:lgr}. This is essentially saying that the saturation or pressure value on the low-fidelity scale is representative in an average sense, but there is important subgrid variability that leads to some inconsistency between quantities at the grid-block scale (100~m) and the instrument scale (cm to m). This variability could be appropriately quantified in terms of converged results for the standard deviations in Table~\ref{tab:lgr}. We cannot claim the $7 \times 7 \times 7$ refinement results represent converged quantities, though the fact that the values in Table~\ref{tab:lgr} are fairly consistent between refinement levels is encouraging.

For saturation, following \cite{sun2019data}, we take the measurement error to be 0.02 (absolute) and the model error to be 0.0865 ($7 \times 7 \times 7$ refinement result). Thus the overall error standard deviation, obtained by adding measurement error and model error variances and taking the square root, is $\sqrt{0.02^2 + 0.0865^2} \approx 0.09$. Note the overall error for saturation is dominated by model error. For pressure, a range of measurement error values has been reported. \citet{sun2012inversion} state that pressure transducer accuracy ranges from 0.1\% to 0.5\% of the measured value. Pressure in our case is around 20~MPa, so measurement error would be in the range of 0.02~MPa to 0.1~MPa. If we take a value near the high end of this range (say 0.095~MPa), and incorporate model error (0.03~MPa), we get an overall error standard deviation of 0.1~MPa. This value, which is consistent with the value applied in \cite{han2023surrogate}, is used in this work. Overall pressure error, in contrast to saturation error, is here dominated by measurement error.

\begin{table}[!ht]
\footnotesize
\renewcommand{\arraystretch}{1.2} 
\centering
\caption{Pressure and saturation bias and standard deviation for different LGR levels.}
\begin{tabular}{cccc}
\hline
     &LGR & Bias & Standard deviation  \\
     \hline
     \multirow{3}{*}{Pressure (MPa)} & $3 \times 3 \times 3$ & 0.0110 & 0.0292\\
     & $5 \times 5 \times 5$ & 0.0137 & 0.0295 \\
     & $7 \times 7 \times 7$ & 0.0168 & 0.0297 \\
     \hline
     \multirow{3}{*}{Saturation} & $3 \times 3 \times 3$ & 0.0091 & 0.0723\\
     & $5 \times 5 \times 5$ & 0.0099 & 0.0763 \\
     & $7 \times 7 \times 7$ & 0.0091 &  0.0865 \\
     \hline
\end{tabular}
\label{tab:lgr}
\end{table}

\subsection{DSI results for true model~1}
We now present history matching results using our new DSI framework. In all history matching examples, the latent variable $\Bxi$ is of dimension 1000. Within ESMDA, we use $N_r = 400$ prior realizations of $\Bxi$. Observed data include pressure and saturation at four time steps (1, 3, 5 and 10~years), measured in layers~10--17 (geological layers~3 and 4), in all eight monitoring wells. The total number of observations is thus 512. Error, which appears in the $C_D$ matrix in Eq.~\ref{eq:esmda_ksi}, is as described above. We assimilate the observation data $N_a = 4$ times. Predictions are then made over the rest of the 20-year injection time frame. 
For each assimilation step, the generation of $N_r = 400$ data realizations using the AAE parameterization requires only 4~seconds on a single Nvidia Tesla A100 GPU. The complete DSI procedure takes about 4~minutes to complete. This timing is dominated by the computation of the inverse of the covariance matrix. 

\begin{figure}[!hbt]
\centering
\begin{minipage}{.45\linewidth}\centering
\includegraphics[trim = 0 0 0 0, clip, width=\linewidth]{ 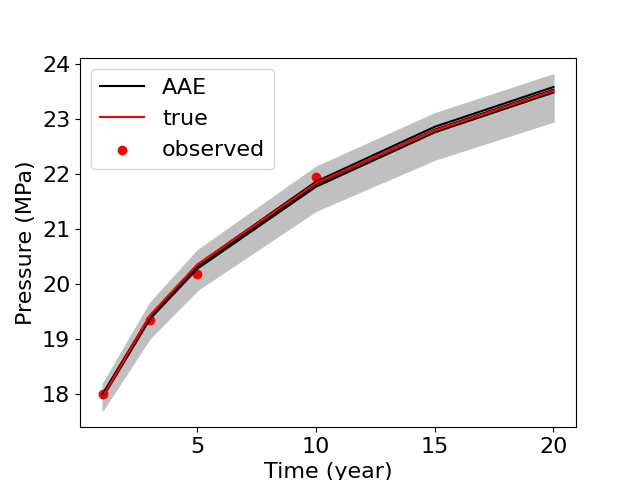}
\subcaption{O7 pressure for layer~15}
\end{minipage}
\begin{minipage}{.45\linewidth}\centering
\includegraphics[trim = 0 0 0 0, clip, width=\linewidth]{ 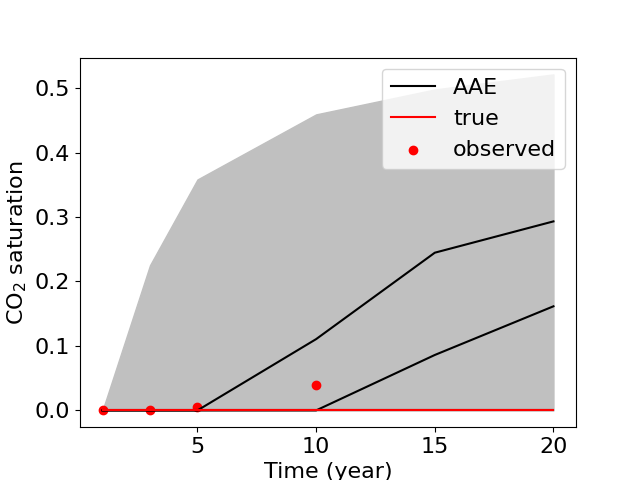}
\subcaption{O7 saturation for layer~15}
\end{minipage}

\begin{minipage}{.45\linewidth}\centering
\includegraphics[trim = 0 0 0 0, clip, width=\linewidth]{ 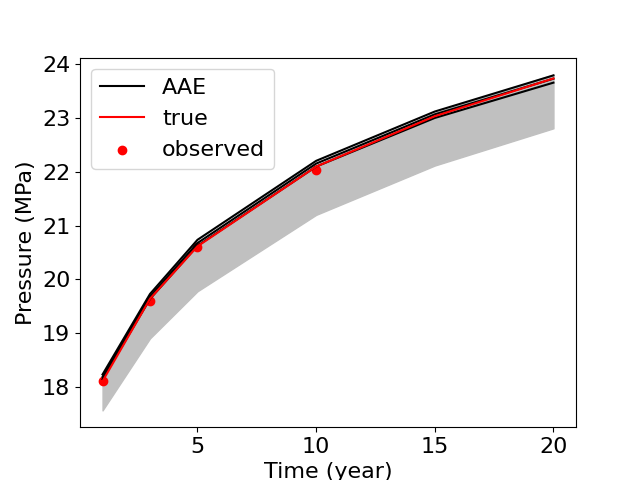}
\subcaption{O8 pressure for layer~10}
\end{minipage}
\begin{minipage}{.45\linewidth}\centering
\includegraphics[trim = 0 0 0 0, clip, width=\linewidth]{ 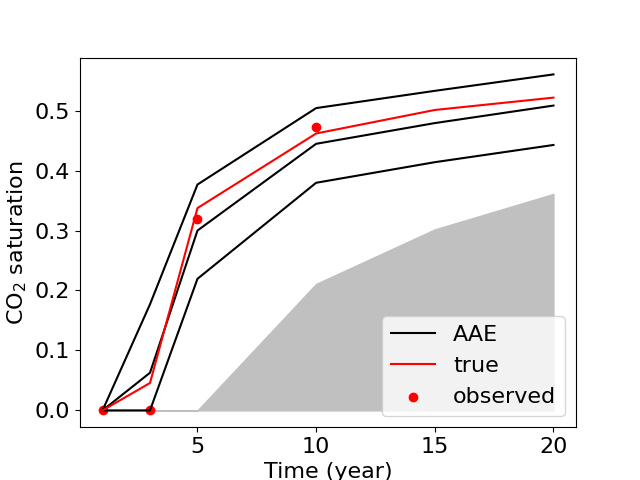}
\subcaption{O8 saturation for layer~10}
\end{minipage}

\caption{Posterior statistics for DSI results with AAE (black curves) for true model~1. Gray shaded areas represent the prior P$_{10}$--P$_{90}$ range, red circles denote the observed data, and red lines denote the true response. Lower, middle and upper curves are P$_{10}$, P$_{50}$ and P$_{90}$ posterior flow responses.} \label{fig:post_flow_obs_0}
\end{figure}

We first consider true model~1, shown in Fig.~\ref{fig:perm_prior}a. Figure~\ref{fig:post_flow_obs_0} presents the prior and posterior statistics for pressure and saturation for layer~15 at monitoring well~O7 and for layer~10 at monitoring well~O8. The gray-shaded regions in the subplots present the P$_{10}$--P$_{90}$ range of the prior statistics. The red curves show the true data, which derive from simulation of the true model (with no error included). The red circles represent the observed data, which include random error. The black lines indicate the P$_{10}$ (lower), P$_{50}$ (middle), and P$_{90}$ (upper) posterior results. 

Significant uncertainty reduction is evident in the pressure data, i.e., the P$_{10}$--P$_{90}$ posterior range is much more narrow than the prior range. For this model the saturation in layer~15 at O7 is close to 0. The posterior predictions capture this behavior quite well. Interesting behavior is observed for saturation in layer~10 at O8 (Fig.~\ref{fig:post_flow_obs_0}d). The data in this case are significantly above the P$_{10}$--P$_{90}$ prior range. DSI posterior predictions nonetheless bracket the data and capture the general response. 

The primary variables, i.e., those provided directly by the DSI procedure, are the  pressure and saturation in each grid block at a set of time steps. Derived quantities, such as average saturation in a layer, are quantities computed from the primary variables. As discussed in \cite{jiang2021deep}, derived quantities may not be captured accurately if the correlations between the primary variables are not properly preserved by the parameterization procedure.

\begin{figure}[!hbt]
\centering
\begin{minipage}{.45\linewidth}\centering
\includegraphics[trim = 0 0 0 0, clip, width=\linewidth]{ 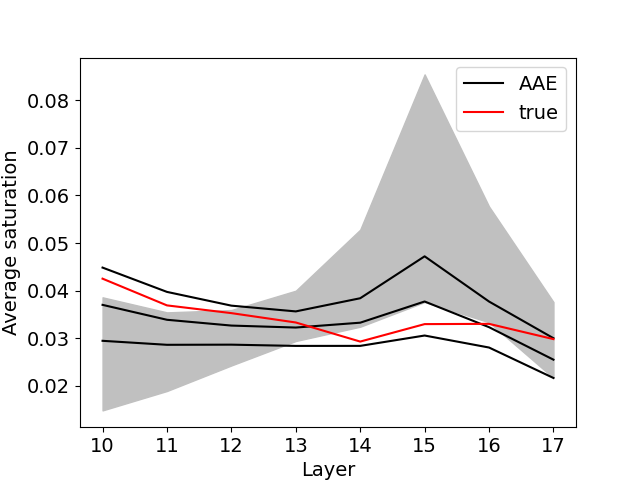}
\subcaption{Average saturation}
\end{minipage}
\begin{minipage}{.45\linewidth}\centering
\includegraphics[trim = 0 0 0 0, clip, width=\linewidth]{ 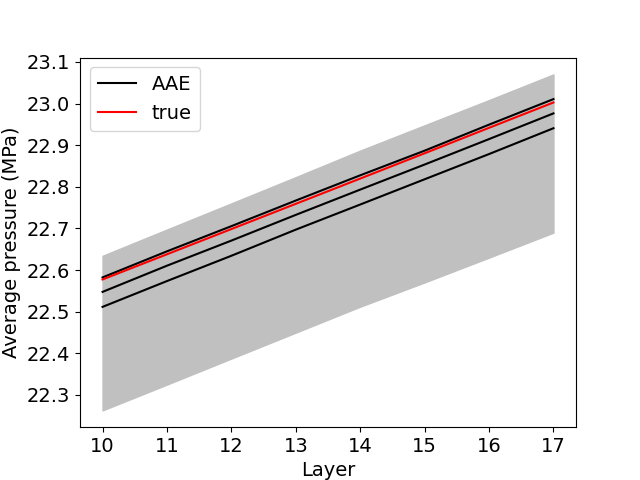}
\subcaption{Average pressure}
\end{minipage}

\caption{Prior and posterior statistics for average saturation and pressure, by layer, for true model~1. Gray shaded areas represent the prior P$_{10}$--P$_{90}$ range and red lines denote the true response. Lower, middle and upper black curves are P$_{10}$, P$_{50}$ and P$_{90}$ posterior results.}\label{fig:post_derived}
\end{figure}

Figure~\ref{fig:post_derived} displays results for key derived quantities -- the predicted average saturation and pressure in layers~10--17 at 20~years. The gray region again indicates the P$_{10}$--P$_{90}$ prior, and the curves are as explained earlier. In the prior, the average saturation displays a clear maximum in layer~15. This large peak does not occur for the true model (red curve), however. We see that DSI is able to capture the correct behavior, with the true response falling within the P$_{10}$--P$_{90}$ posterior range. This is the case even though a portion of the true response is outside the P$_{10}$--P$_{90}$ prior range. Average layer pressure (Fig.~\ref{fig:post_derived}b) increases linearly with depth, as expected. Significant uncertainty reduction, and accurate DSI predictions, are again observed.

We now present prior and posterior pressure and saturation fields at 20~years. Because we have 400 prior and posterior $\Bxi$ vectors, each corresponding to different pressure and saturation fields, it is important to use a systematic procedure to select representative fields. To accomplish this, we apply a two-step clustering method. Specifically, for prior pressure fields, a K-means procedure is applied to classify the 400 fields into five clusters. A K-medoids method is then used to select the cluster centers of these five clusters, which are then taken to be representative of the 400 prior pressure fields. This procedure is repeated to provide posterior pressure fields and prior and posterior saturation fields.

\begin{figure}[!htb]
\centering
\begin{minipage}{.19\linewidth}\centering
\includegraphics[trim = 0 0 0 0, clip, width=\linewidth]{ 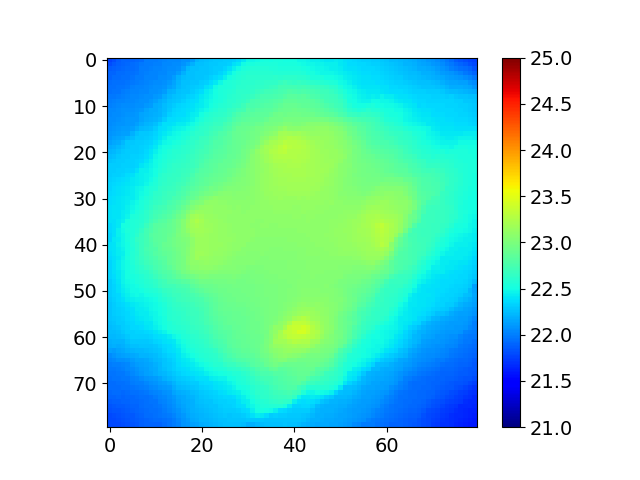}
\subcaption{Realization~1}
\end{minipage}
\begin{minipage}{.19\linewidth}\centering
\includegraphics[trim = 0 0 0 0, clip, width=\linewidth]{ 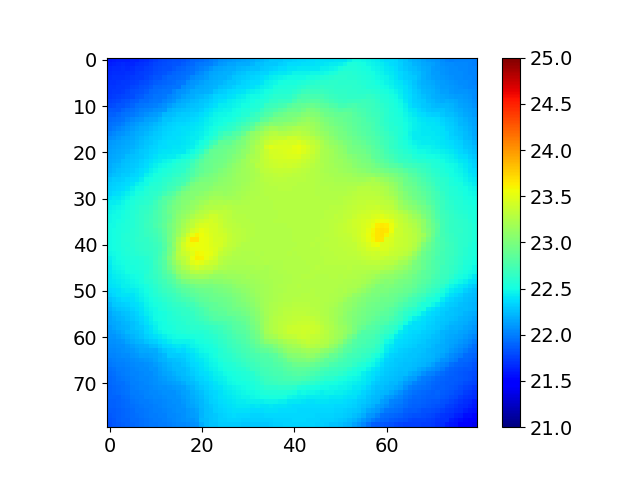}
\subcaption{Realization~2}
\end{minipage}
\begin{minipage}{.19\linewidth}\centering
\includegraphics[trim = 0 0 0 0, clip, width=\linewidth]{ 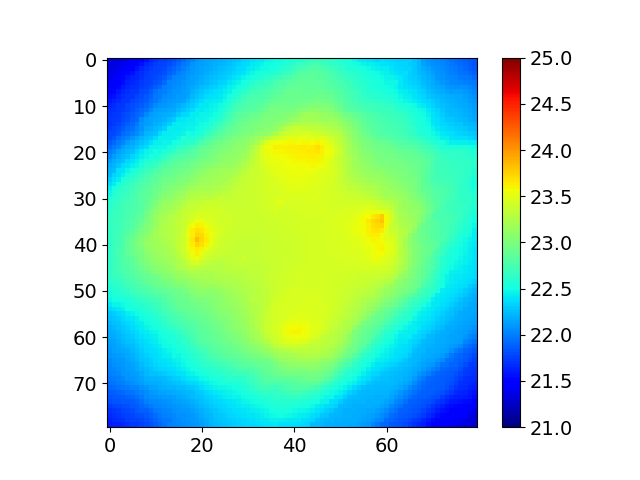}
\subcaption{Realization~3}
\end{minipage}
\begin{minipage}{.19\linewidth}\centering
\includegraphics[trim = 0 0 0 0, clip, width=\linewidth]{ 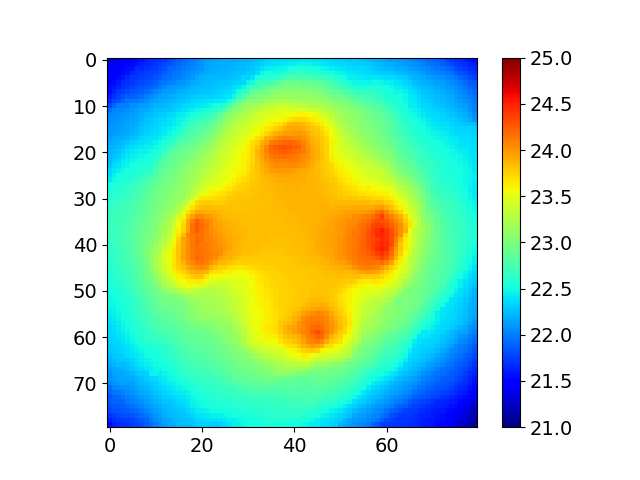}
\subcaption{Realization~4}
\end{minipage}
\begin{minipage}{.19\linewidth}\centering
\includegraphics[trim = 0 0 0 0, clip, width=\linewidth]{ 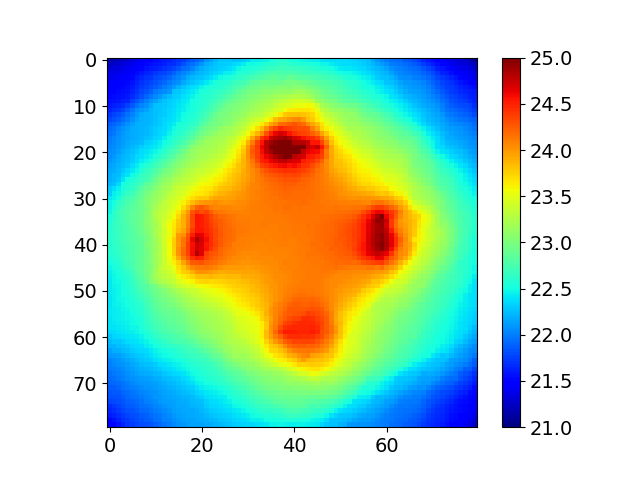}
\subcaption{Realization~5}
\end{minipage}
\begin{minipage}{.19\linewidth}\centering
\includegraphics[trim = 0 0 0 0, clip, width=\linewidth]{ 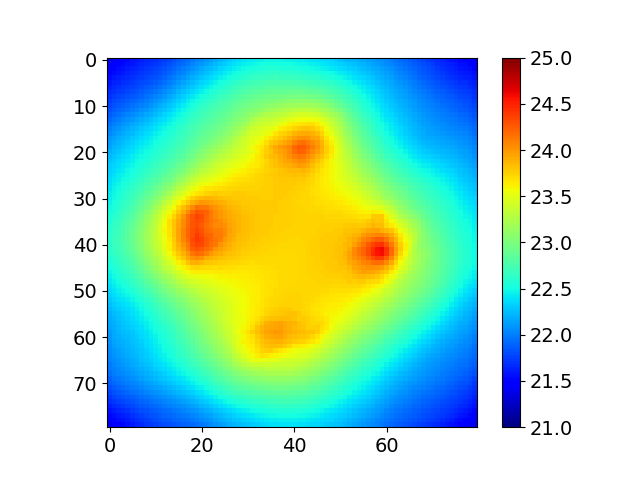}
\subcaption{Realization~1}
\end{minipage}
\begin{minipage}{.19\linewidth}\centering
\includegraphics[trim = 0 0 0 0, clip, width=\linewidth]{ 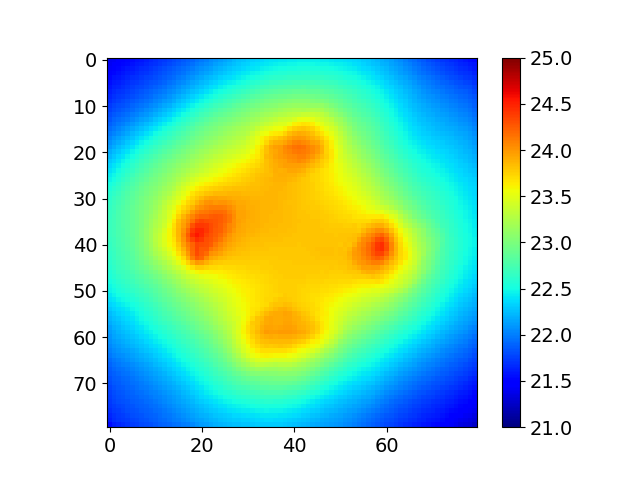}
\subcaption{Realization~2}
\end{minipage}
\begin{minipage}{.19\linewidth}\centering
\includegraphics[trim = 0 0 0 0, clip, width=\linewidth]{ 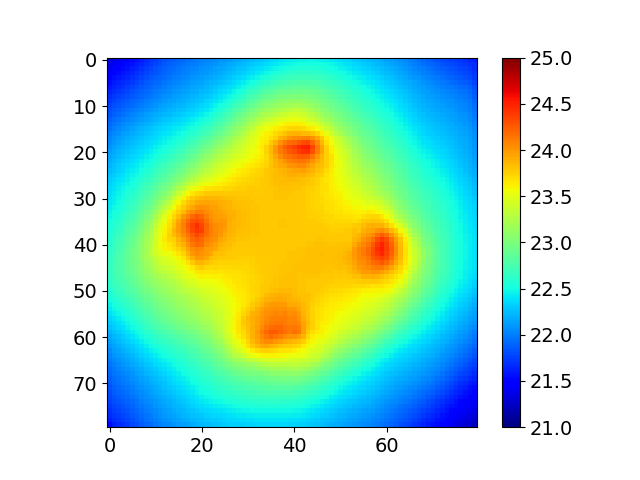}
\subcaption{Realization~3}
\end{minipage}
\begin{minipage}{.19\linewidth}\centering
\includegraphics[trim = 0 0 0 0, clip, width=\linewidth]{ 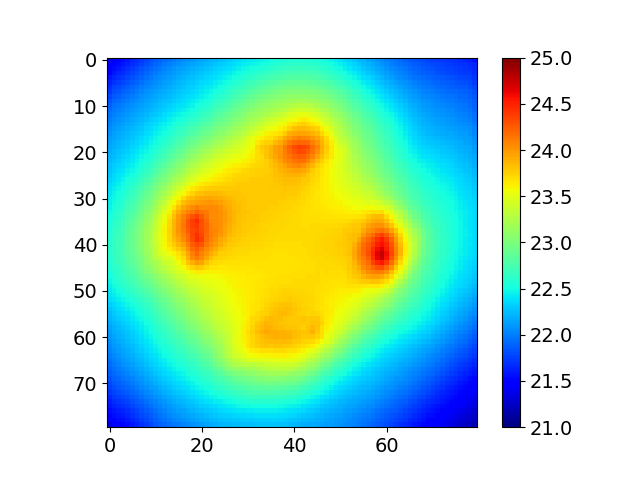}
\subcaption{Realization~4}
\end{minipage}
\begin{minipage}{.19\linewidth}\centering
\includegraphics[trim = 0 0 0 0, clip, width=\linewidth]{ 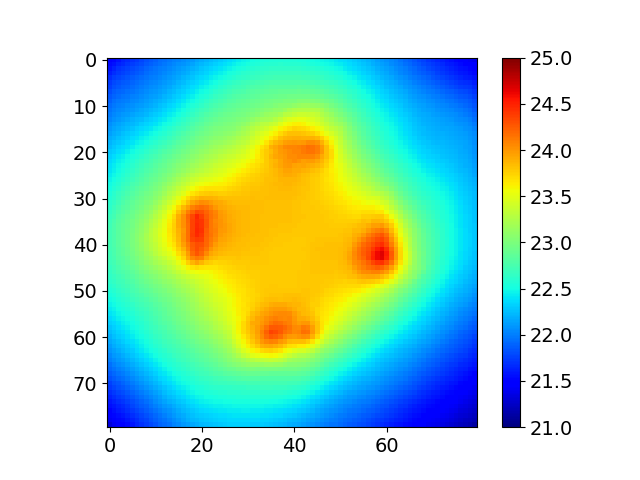}
\subcaption{Realization~5}
\end{minipage}
\caption{Representative prior (upper row) and DSI posterior (lower row) pressure fields for true model~1 for layer~15 at 20~years. True pressure field for this case shown in Fig.~\ref{fig:prior_p_map}a.}\label{fig:post_p_map}
\end{figure}

Figure~\ref{fig:post_p_map} displays the five representative prior (upper row) and posterior (lower row) pressure fields for layer~15 at 20~years. The pressure field corresponding to true model~1 is shown in Fig.~\ref{fig:prior_p_map}a. The prior fields clearly display substantial variation, with pressure extending over a large range. The variation between the posterior fields is much smaller, and these fields are in close visual agreement with the true solution in Fig.~\ref{fig:prior_p_map}a. This illustrates the uncertainty reduction achieved by DSI with the AAE parameterization. 

\begin{figure}[!htb]
\centering
\begin{minipage}{.19\linewidth}\centering
\includegraphics[trim = 0 0 0 0, clip, width=\linewidth]{ 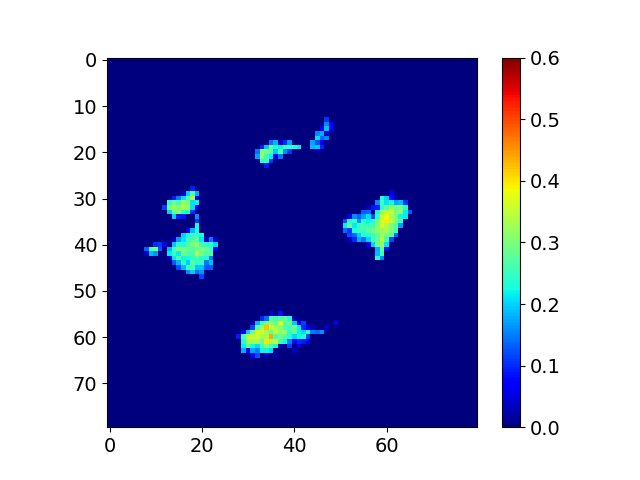}
\subcaption{Realization~1}
\end{minipage}
\begin{minipage}{.19\linewidth}\centering
\includegraphics[trim = 0 0 0 0, clip, width=\linewidth]{ 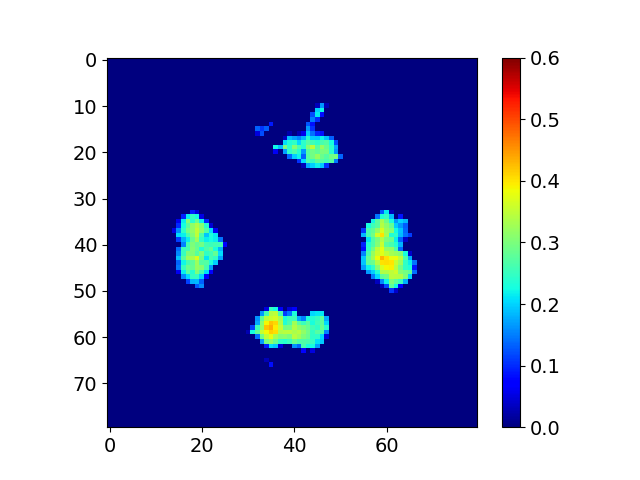}
\subcaption{Realization~2}
\end{minipage}
\begin{minipage}{.19\linewidth}\centering
\includegraphics[trim = 0 0 0 0, clip, width=\linewidth]{ 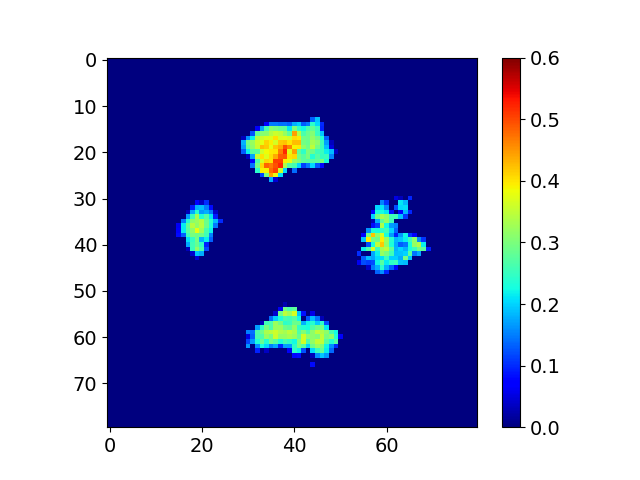}
\subcaption{Realization~3}
\end{minipage}
\begin{minipage}{.19\linewidth}\centering
\includegraphics[trim = 0 0 0 0, clip, width=\linewidth]{ 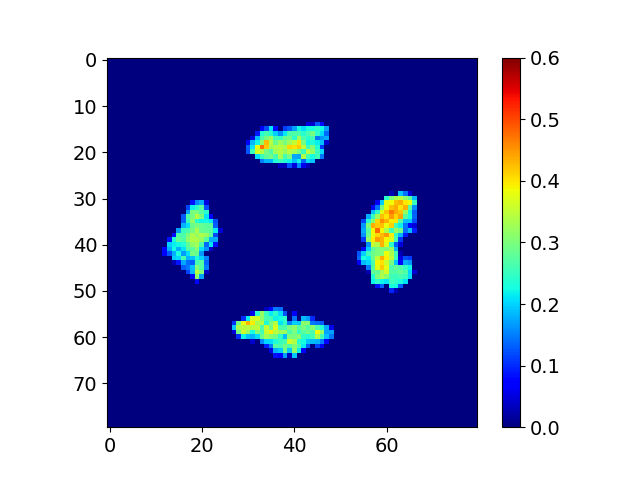}
\subcaption{Realization~4}
\end{minipage}
\begin{minipage}{.19\linewidth}\centering
\includegraphics[trim = 0 0 0 0, clip, width=\linewidth]{ 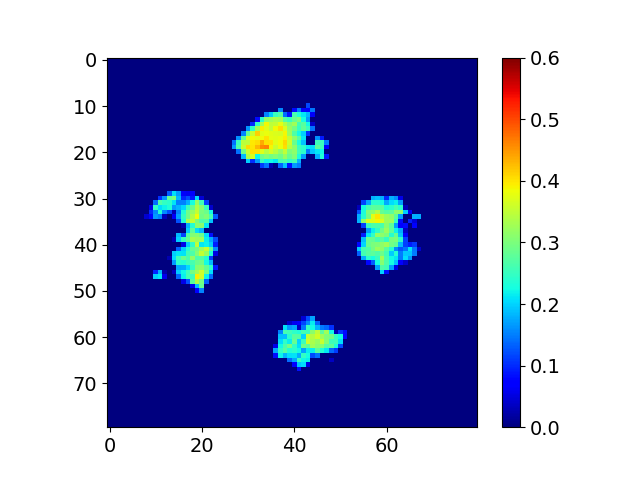}
\subcaption{Realization~5}
\end{minipage}
\begin{minipage}{.19\linewidth}\centering
\includegraphics[trim = 0 0 0 0, clip, width=\linewidth]{ 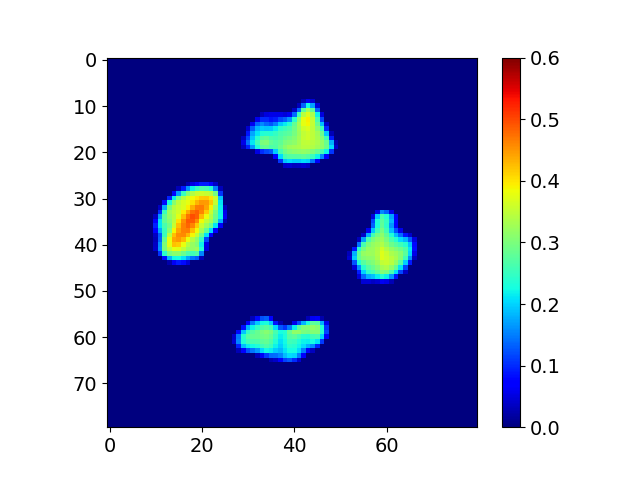}
\subcaption{Realization~1}
\end{minipage}
\begin{minipage}{.19\linewidth}\centering
\includegraphics[trim = 0 0 0 0, clip, width=\linewidth]{ 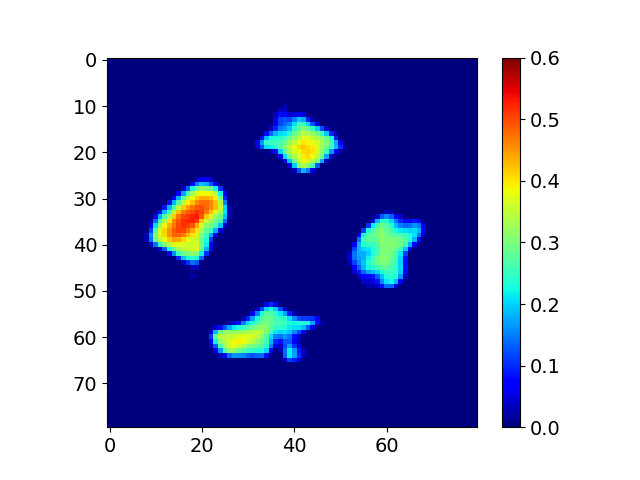}
\subcaption{Realization~2}
\end{minipage}
\begin{minipage}{.19\linewidth}\centering
\includegraphics[trim = 0 0 0 0, clip, width=\linewidth]{ 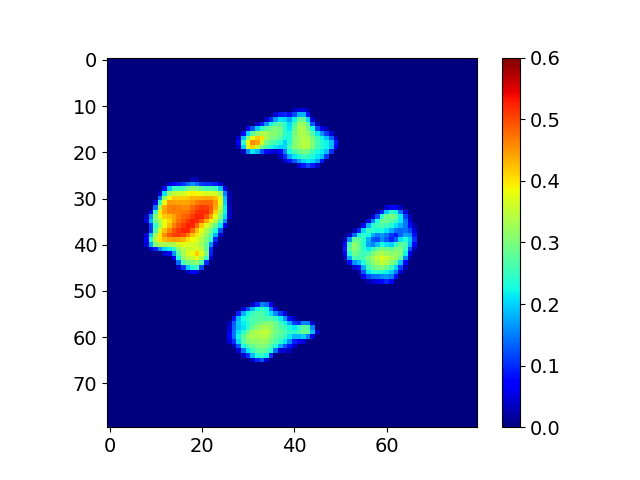}
\subcaption{Realization~3}
\end{minipage}
\begin{minipage}{.19\linewidth}\centering
\includegraphics[trim = 0 0 0 0, clip, width=\linewidth]{ 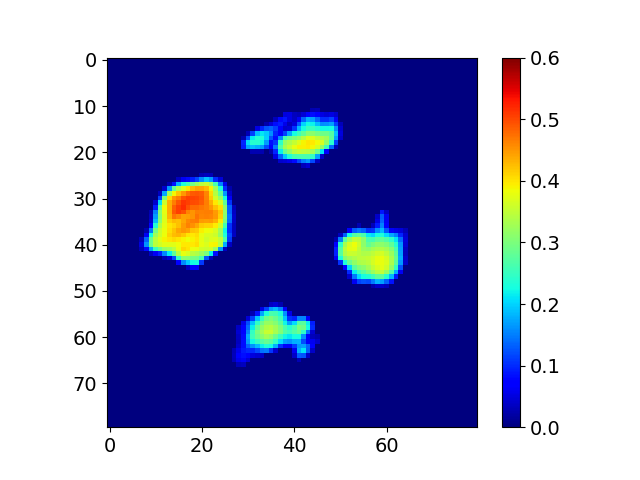}
\subcaption{Realization~4}
\end{minipage}
\begin{minipage}{.19\linewidth}\centering
\includegraphics[trim = 0 0 0 0, clip, width=\linewidth]{ 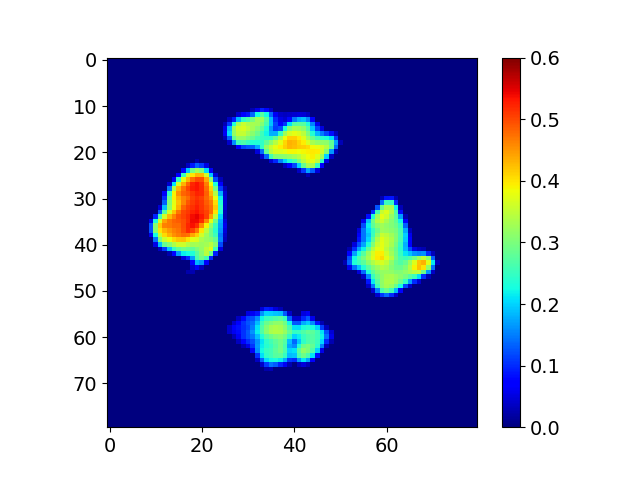}
\subcaption{Realization~5}
\end{minipage}
\caption{Representative prior (upper row) and DSI posterior (lower row) saturation fields for true model~1 for layer~10 at 20~years. True saturation field for this case shown in Fig.~\ref{fig:prior_s_map_1}a.} \label{fig:post_s_map}
\end{figure}

Prior and posterior saturation fields for layer~10 at 20~years are shown in Fig.~\ref{fig:post_s_map}. The corresponding true saturation is shown in Fig.~\ref{fig:prior_s_map_1}a. The representative prior fields display only moderate variability and relatively low CO$_2$ saturations in this layer (consistent with the prior range in Fig.~\ref{fig:post_derived}a). The true CO$_2$ plume on the left is much larger than in the representative prior realizations. This outlier behavior is, however, clearly captured by DSI, as is evident in the representative posterior realizations in Fig.~\ref{fig:post_s_map}.
 
Because the DSI methodology provides posterior predictions without the need for any additional flow simulations or AAE training, it is very efficient for exploring the impact of data types, data error, etc. on predictions. This is important because error can be challenging to quantify, and the sensitivity of predictions to error specification may be of interest. Here we consider the effect of total error on pressure and saturation predictions at 20 years. For this evaluation, we compute the cell-by-cell difference in pressure and saturation between each DSI posterior field and true model~1 (and similarly for the prior fields). An overall mean absolute difference for each pressure and saturation realization is then computed. These differences quantify the discrepancy between (prior or posterior) samples and the true solution, and illustrate the level of uncertainty reduction achieved.

\begin{figure}[!hbt]
\centering
\begin{minipage}{.7\linewidth}\centering
\includegraphics[trim = 0 0 0 0, clip, width=\linewidth]{ 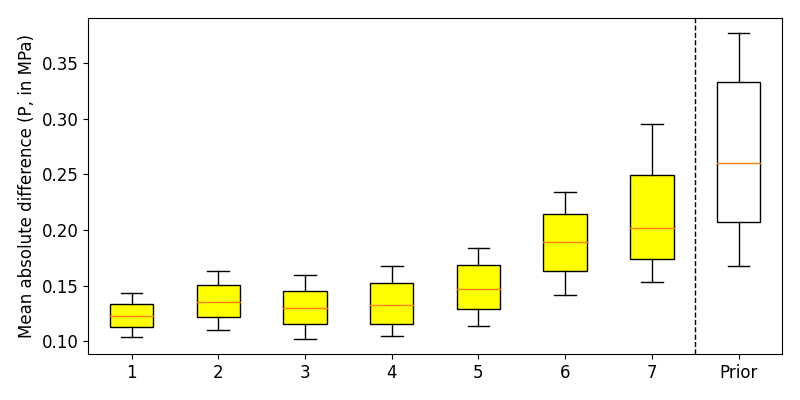}
\subcaption{Pressure (MPa)}
\end{minipage}
\begin{minipage}{.7\linewidth}\centering
\includegraphics[trim = 0 0 0 0, clip, width=\linewidth]{ 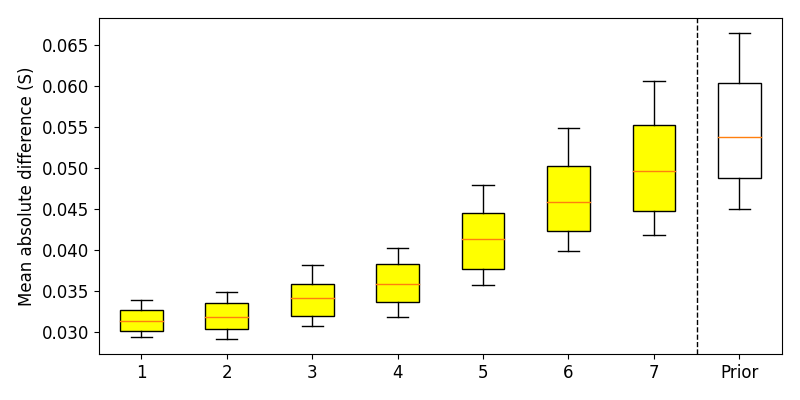}
\subcaption{Saturation}
\end{minipage}
\caption{Mean absolute differences in pressure and saturation between true model~1 and prior fields (white boxes), and between true model~1 and posterior fields (yellow boxes) for a range of error specifications. See text for detailed error settings and box-plot quantities.}\label{fig:post_mae}
\end{figure}

Figure~\ref{fig:post_mae} displays box plots of the mean absolute differences for prior and posterior pressure and saturation fields. For the posterior fields, seven different error specifications are considered. In these and subsequent box plots, the top and bottom of each box show the P$_{75}$ and P$_{25}$ (75th and 25th percentile) mean absolute differences, and the solid red line provides the P$_{50}$ result. The lines extending beyond the boxes show the P$_{90}$ and P$_{10}$ errors.

The white boxes in Fig.~\ref{fig:post_mae} show results for the prior simulations. The prior results display a wide range of responses and the largest differences with the true results. The yellow boxes show the differences for posterior samples for a range of total error settings. Specifically, we specify the error for pressure and saturation as 0.02~MPa and 0.02 for case~1 (denoted by `1' on the plots), 0.05~MPa and 0.05 for case~2, 0.1~MPa and 0.09 for case 3 (these values were used in the results shown above), 0.15~MPa and 0.15 for case~4, 0.5~MPa and 0.5 for case~5, 1~MPa and 0.5 for case~6, 2~MPa and 1 for case~7. Error increases as we move through the cases, and the box plots reflect the impact of the error specification on posterior results. As would be expected, posterior uncertainty increases (i.e., the amount of uncertainty reduction from history matching decreases) as error is increased. In fact, with the largest error specification (case~7), the results are relatively close to the prior, since the observations in this case are not very informative. We also see that, while the saturation differences consistently increase with increasing error, the pressure differences show relatively little variation over error cases~1--4. Assessments of this type can be applied to quickly evaluate the impact of additional measurements and/or different monitoring well configurations.

\subsection{DSI results for true model~2}
\begin{figure}[!hbt]
\centering
\begin{minipage}{.45\linewidth}\centering
\includegraphics[trim = 0 0 0 0, clip, width=\linewidth]{ 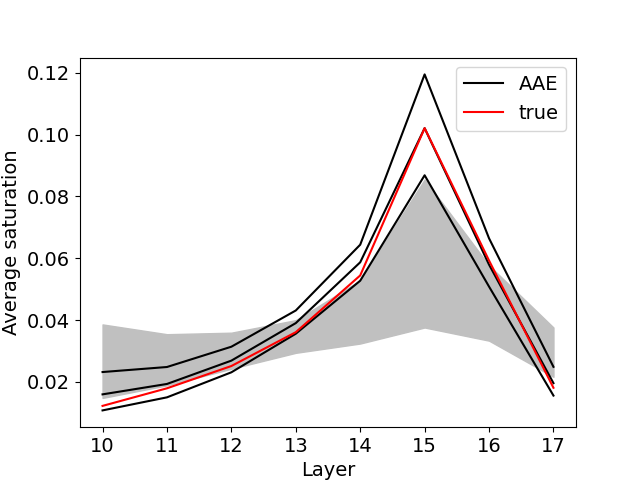}
\subcaption{Average saturation}
\end{minipage}
\begin{minipage}{.45\linewidth}\centering
\includegraphics[trim = 0 0 0 0, clip, width=\linewidth]{ 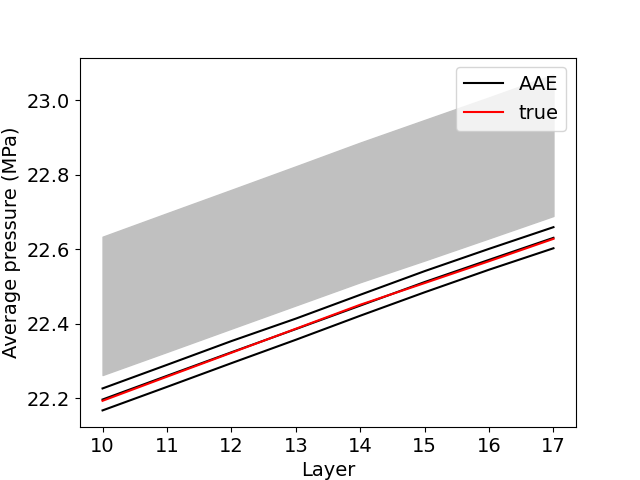}
\subcaption{Average pressure}
\end{minipage}

\caption{Prior and posterior statistics for average saturation and pressure, by layer, for true model~2. Gray shaded areas represent the prior P$_{10}$--P$_{90}$ range and red lines denote the true response. Lower, middle and upper black curves are P$_{10}$, P$_{50}$ and P$_{90}$ posterior results.}\label{fig:post_derived_1}
\end{figure}

The log-permeability field for true model~2 is shown in Fig.~\ref{fig:perm_prior}b. Prior and posterior results for layer-average saturation and pressure, for layers~10--17, are presented in Fig.~\ref{fig:post_derived_1}. In this case the highest average saturation for the true model is observed in layer~10 (the maximum for true model~1 was in layer~15). The true saturation results are beyond the P$_{90}$ prior values in several of the layers, including layer~15, though the DSI posterior predictions clearly track the true-model results. In the case of pressure, the true results are consistently below the P$_{10}$ prior value, though again DSI provides accurate results and significant uncertainty reduction.

\begin{figure}[!htb]
\centering
\begin{minipage}{.19\linewidth}\centering
\includegraphics[trim = 0 0 0 0, clip, width=\linewidth]{ 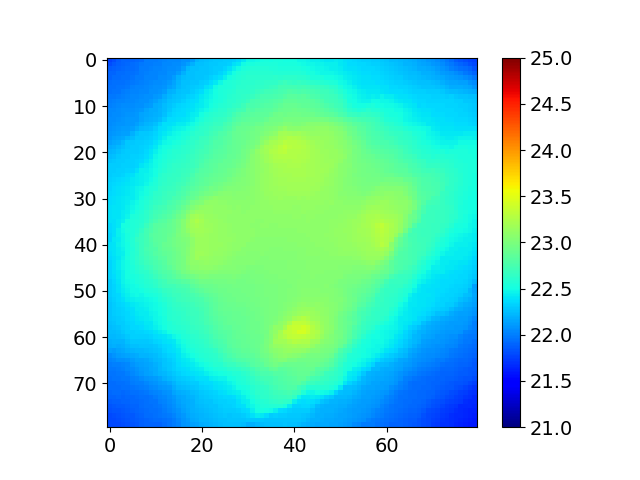}
\subcaption{Realization~1}
\end{minipage}
\begin{minipage}{.19\linewidth}\centering
\includegraphics[trim = 0 0 0 0, clip, width=\linewidth]{ 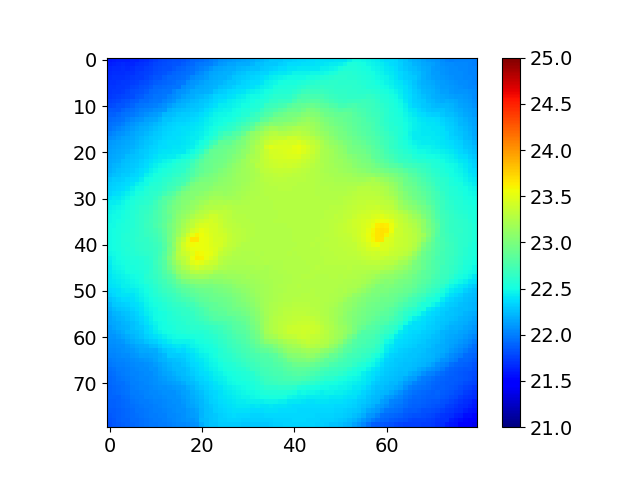}
\subcaption{Realization~2}
\end{minipage}
\begin{minipage}{.19\linewidth}\centering
\includegraphics[trim = 0 0 0 0, clip, width=\linewidth]{ 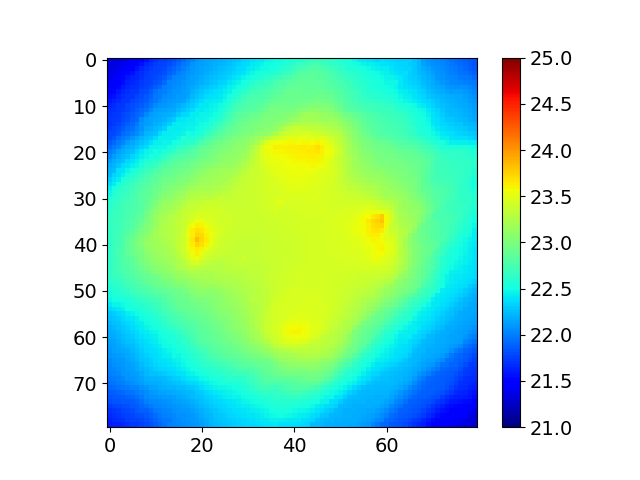}
\subcaption{Realization~3}
\end{minipage}
\begin{minipage}{.19\linewidth}\centering
\includegraphics[trim = 0 0 0 0, clip, width=\linewidth]{ 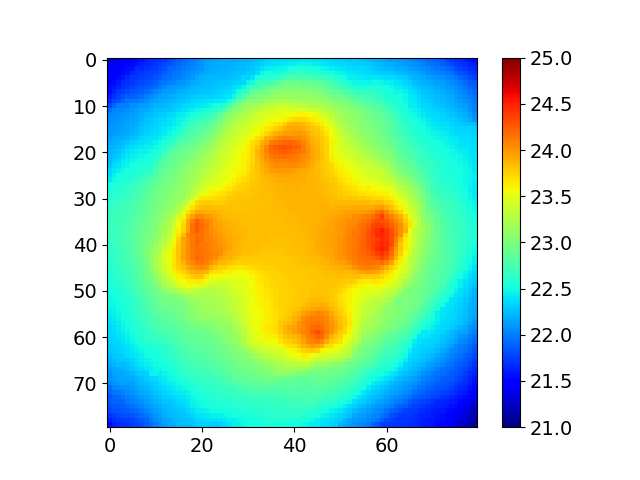}
\subcaption{Realization~4}
\end{minipage}
\begin{minipage}{.19\linewidth}\centering
\includegraphics[trim = 0 0 0 0, clip, width=\linewidth]{ 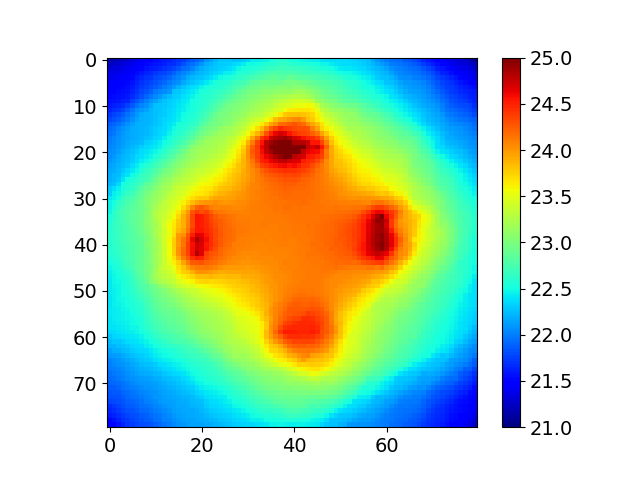}
\subcaption{Realization~5}
\end{minipage}
\begin{minipage}{.19\linewidth}\centering
\includegraphics[trim = 0 0 0 0, clip, width=\linewidth]{ 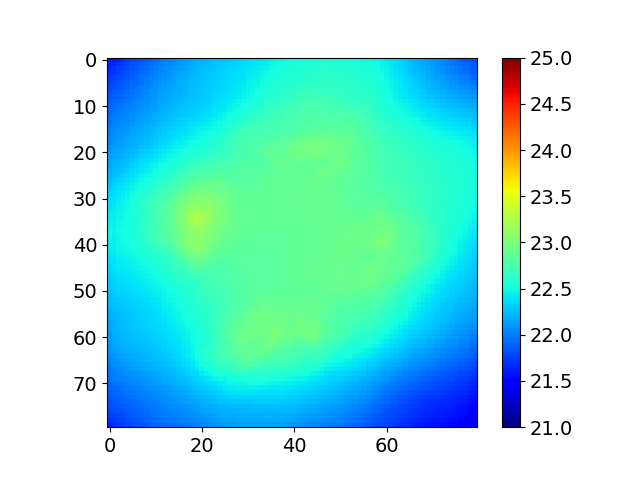}
\subcaption{Realization~1}
\end{minipage}
\begin{minipage}{.19\linewidth}\centering
\includegraphics[trim = 0 0 0 0, clip, width=\linewidth]{ 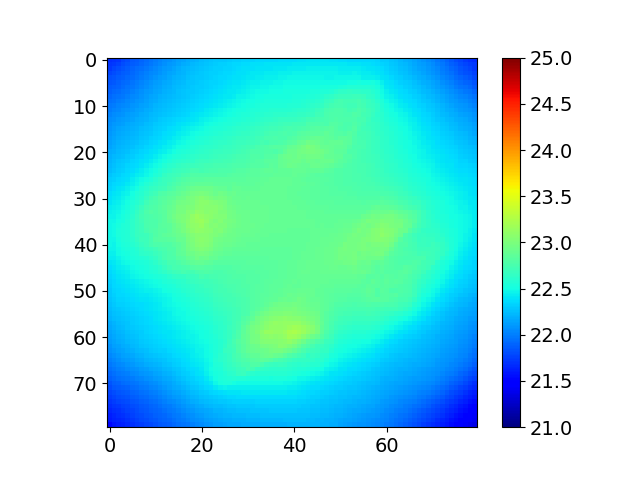}
\subcaption{Realization~2}
\end{minipage}
\begin{minipage}{.19\linewidth}\centering
\includegraphics[trim = 0 0 0 0, clip, width=\linewidth]{ 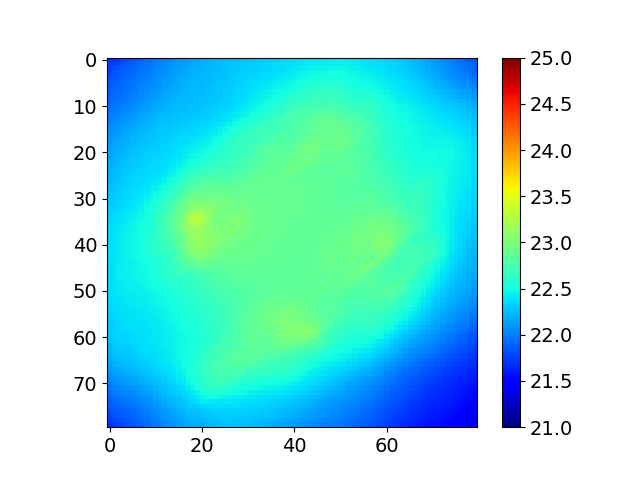}
\subcaption{Realization~3}
\end{minipage}
\begin{minipage}{.19\linewidth}\centering
\includegraphics[trim = 0 0 0 0, clip, width=\linewidth]{ 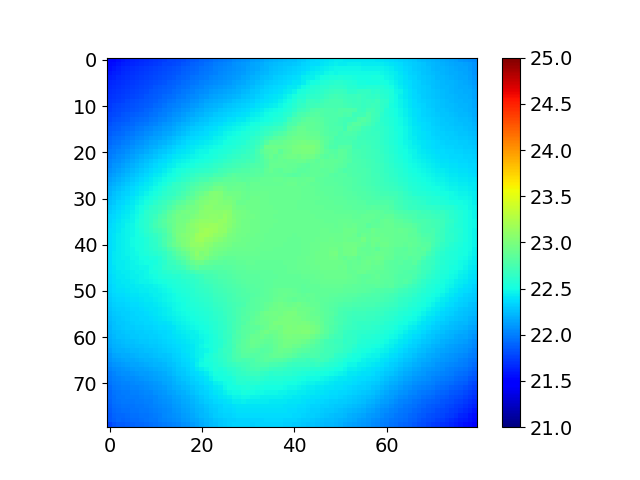}
\subcaption{Realization~4}
\end{minipage}
\begin{minipage}{.19\linewidth}\centering
\includegraphics[trim = 0 0 0 0, clip, width=\linewidth]{ 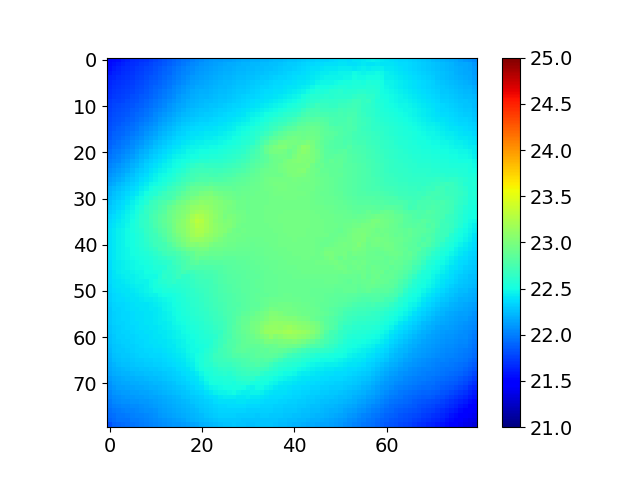}
\subcaption{Realization~5}
\end{minipage}
\caption{Representative prior (upper row) and DSI posterior (lower row) pressure fields for true model~2 for layer~15 at 20~years. True pressure field for this case shown in Fig.~\ref{fig:prior_p_map}b.} \label{fig:post_p_map_1}
\end{figure}

\begin{figure}[!htb]
\centering
\begin{minipage}{.19\linewidth}\centering
\includegraphics[trim = 0 0 0 0, clip, width=\linewidth]{ 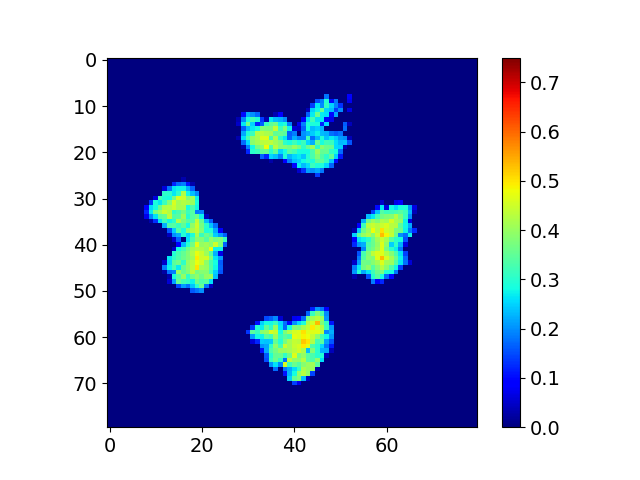}
\subcaption{Realization~1}
\end{minipage}
\begin{minipage}{.19\linewidth}\centering
\includegraphics[trim = 0 0 0 0, clip, width=\linewidth]{ 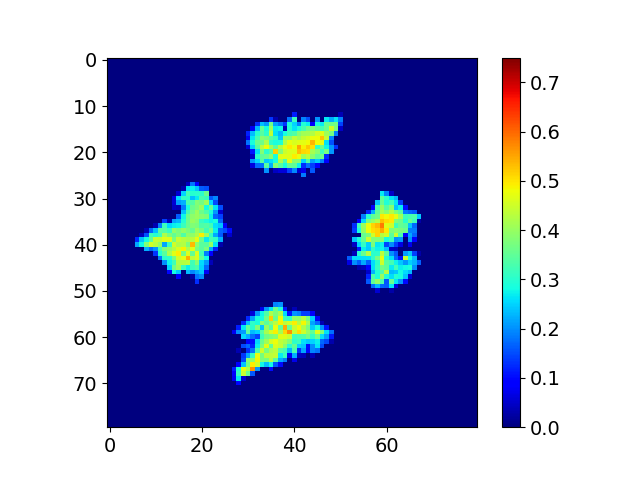}
\subcaption{Realization~2}
\end{minipage}
\begin{minipage}{.19\linewidth}\centering
\includegraphics[trim = 0 0 0 0, clip, width=\linewidth]{ 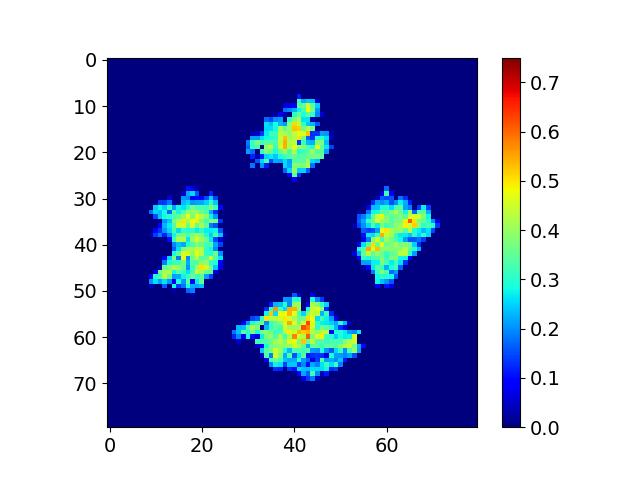}
\subcaption{Realization~3}
\end{minipage}
\begin{minipage}{.19\linewidth}\centering
\includegraphics[trim = 0 0 0 0, clip, width=\linewidth]{ 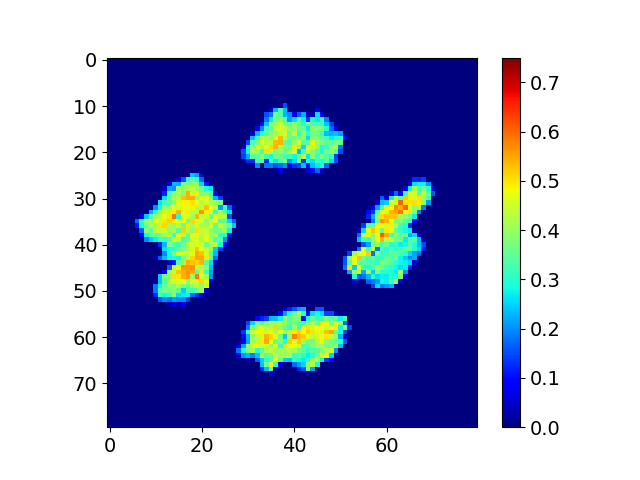}
\subcaption{Realization~4}
\end{minipage}
\begin{minipage}{.19\linewidth}\centering
\includegraphics[trim = 0 0 0 0, clip, width=\linewidth]{ 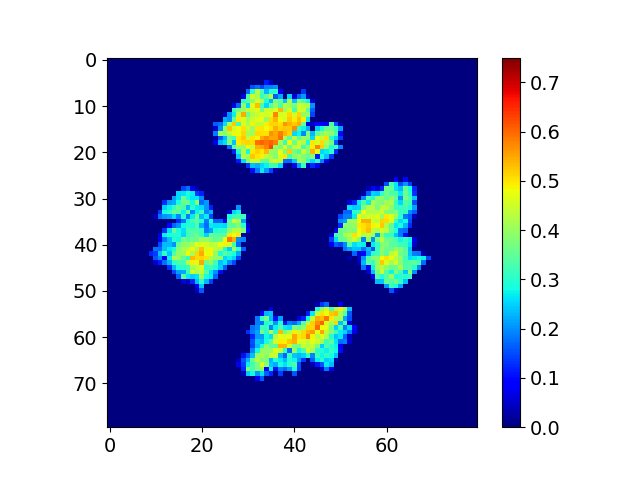}
\subcaption{Realization~5}
\end{minipage}
\begin{minipage}{.19\linewidth}\centering
\includegraphics[trim = 0 0 0 0, clip, width=\linewidth]{ 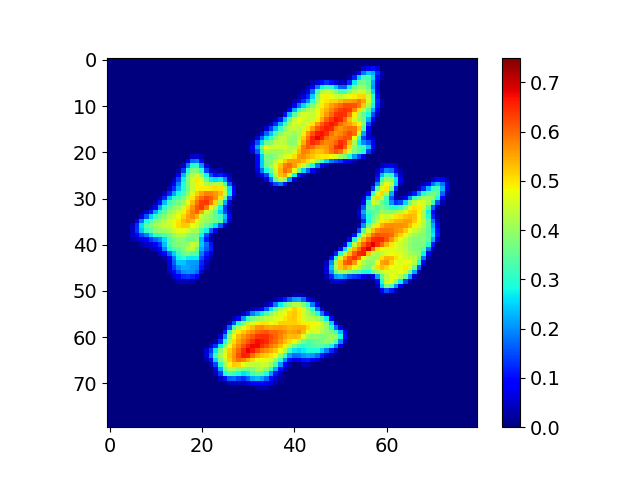}
\subcaption{Realization~1}
\end{minipage}
\begin{minipage}{.19\linewidth}\centering
\includegraphics[trim = 0 0 0 0, clip, width=\linewidth]{ 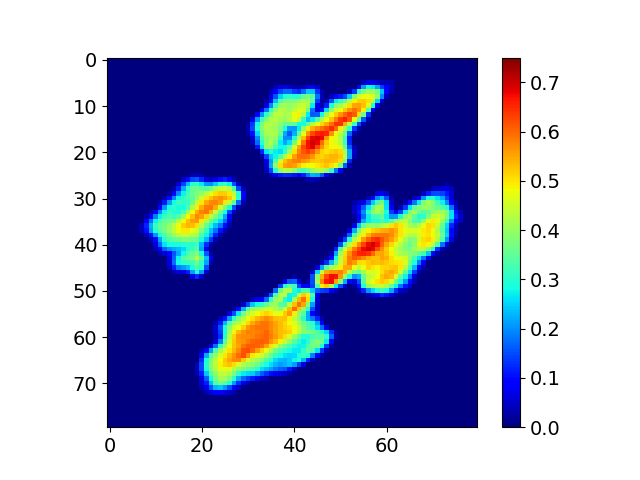}
\subcaption{Realization~2}
\end{minipage}
\begin{minipage}{.19\linewidth}\centering
\includegraphics[trim = 0 0 0 0, clip, width=\linewidth]{ 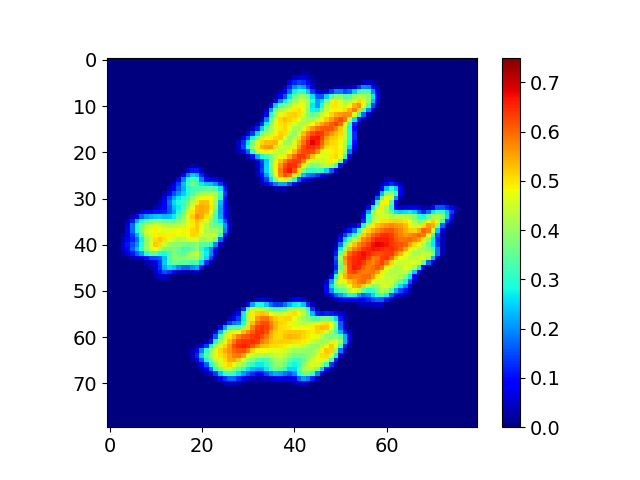}
\subcaption{Realization~3}
\end{minipage}
\begin{minipage}{.19\linewidth}\centering
\includegraphics[trim = 0 0 0 0, clip, width=\linewidth]{ 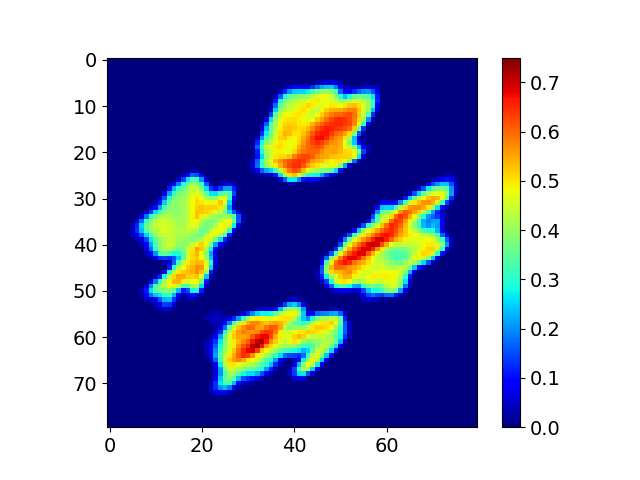}
\subcaption{Realization~4}
\end{minipage}
\begin{minipage}{.19\linewidth}\centering
\includegraphics[trim = 0 0 0 0, clip, width=\linewidth]{ 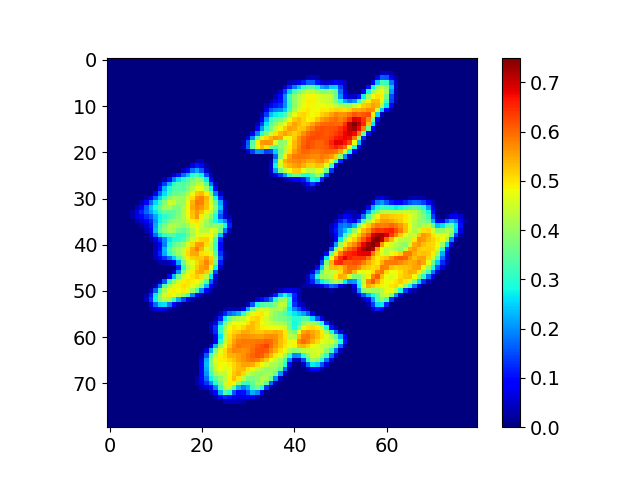}
\subcaption{Realization~5}
\end{minipage}
\caption{Representative prior (upper row) and DSI posterior (lower row) saturation fields for true model~2 for layer~15 at 20~years. True saturation field for this case shown in Fig.~\ref{fig:prior_s_map_1}b.} \label{fig:post_s_map_1}
\end{figure}

Representative prior and posterior pressure and saturation fields are again identified using the K-means and K-medoids procedure described earlier. The resulting fields are shown in Figs.~\ref{fig:post_p_map_1} and \ref{fig:post_s_map_1}. The pressure and saturation fields for true model~2 appear in Figs.~\ref{fig:prior_p_map}b and \ref{fig:prior_s_map_2}b. The pressure in layer~15 is relatively low, and this is captured in the posterior fields, which again show much less variability than the prior fields. The true saturation field in Fig.~\ref{fig:prior_s_map_2}b exhibits large plumes oriented at roughly 45$^\circ$. This general behavior is evident in the posterior fields in Fig.~\ref{fig:post_s_map_1}.

\subsection{DSI results for multiple true models}
In our final assessment, we further evaluate DSI performance, in terms of summary results, for five true models (two of these are true models~1 and 2). We compare DSI posterior results generated using the AAE parameterization with those from a simpler treatment. The latter entails the use of principal component analysis (PCA) followed by histogram transformation (HT). This PCA+HT approach, implemented for carbon storage problems in \cite{sun2019data}, captures the marginal distribution of data variables but may not accurately represent correlations between different data variables. As shown by \citet{jiang2021data}, this can lead to the overestimation of posterior uncertainty in some cases.

\begin{figure}[htbp!]
\centering
\begin{minipage}{0.75\linewidth}\centering
\includegraphics[trim = 0 0 0 0, clip, width=\linewidth]{ 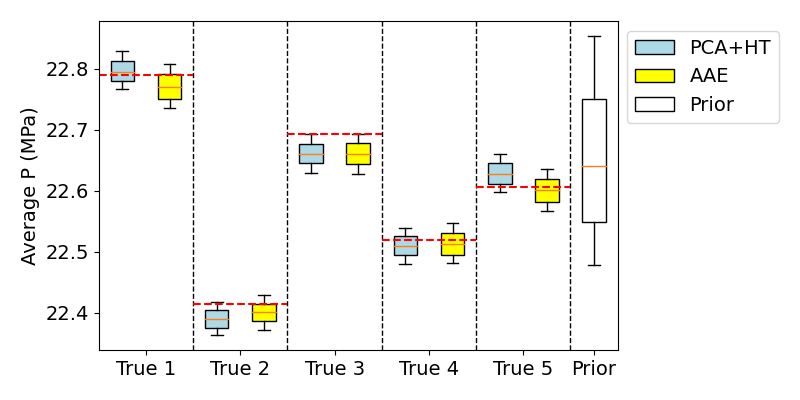}
\subcaption{Average pressure}
\end{minipage}
\begin{minipage}{0.75\linewidth}\centering
\includegraphics[trim = 0 0 0 0, clip, width=\linewidth]{ 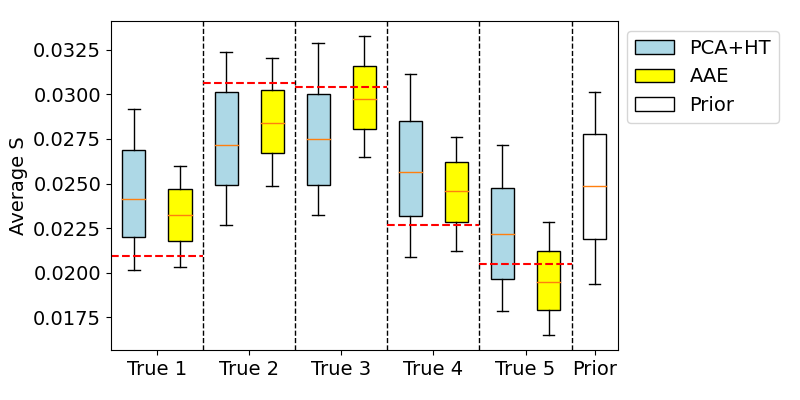}
\subcaption{Average saturation}
\end{minipage}
\caption{Prior and posterior results for average pressure and saturation over layers~10-17 at 20~years for five true models. Posterior DSI results shown for PCA+HT and AAE parameterizations. True results indicated by the red dashed lines.} \label{fig:multi_qoi}
\end{figure}

Figure~\ref{fig:multi_qoi} presents box plots for posterior predictions of two overall (derived) quantities. Results are shown for average pressure and average saturation, with these averages computed over all cells in layers~10--17 at year 20. The three additional true models (3--5) are three new random realizations not included in the prior ensemble, though their metaparameters are within the prior ranges. The white boxes in Fig.~\ref{fig:multi_qoi} show prior results for average pressure and saturation. The yellow boxes correspond to posterior results generated using the AAE parameterization, and the blue boxes to posterior results using PCA+HT. The true results are displayed as the red dashed lines. It is apparent that results for the five true models fall in different regions of the prior. There is general correspondence between the AAE and PCA+HT results, and in all cases the true results fall within the P$_{10}$--P$_{90}$ posterior range. Substantial uncertainty reduction for average pressure is achieved for all true models. For saturation, DSI with AAE provides more uncertainty reduction than PCA+HT.

\begin{figure}[htbp!]
\centering
\begin{minipage}{0.75\linewidth}\centering
\includegraphics[trim = 0 0 0 0, clip, width=\linewidth]{ 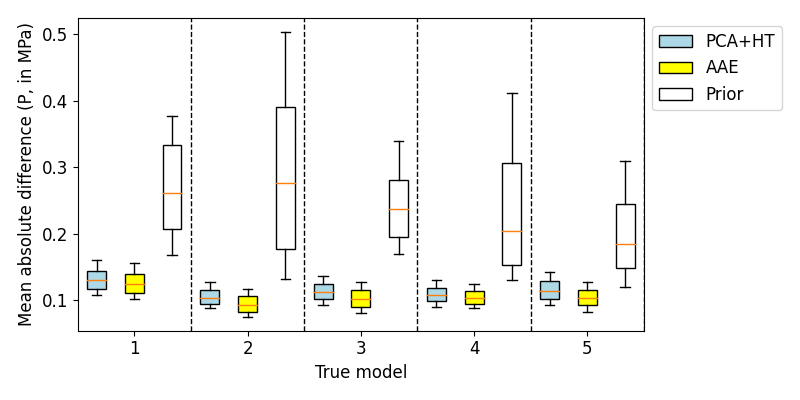}
\subcaption{Pressure}
\end{minipage}
\begin{minipage}{0.75\linewidth}\centering
\includegraphics[trim = 0 0 0 0, clip, width=\linewidth]{ 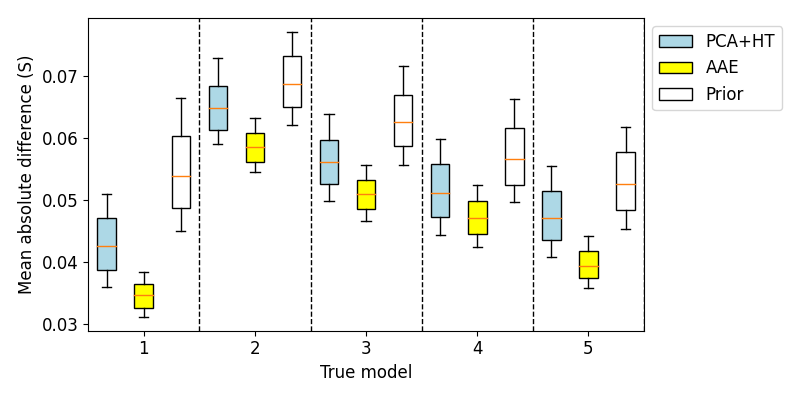}
\subcaption{Saturation}
\end{minipage}
\caption{Mean absolute differences in pressure and saturation between true models and prior fields (white boxes), and between true models and posterior fields  using PCA+HT (blue boxes) and AAE (yellow boxes) parameterizations.}\label{fig:multi_mae}
\end{figure}

We next present, in Fig.~\ref{fig:multi_mae}, mean absolute difference plots for DSI results using the two parameterizations. These quantities are calculated the same way as in Fig.~\ref{fig:post_mae}. Prior results are also shown, so the amount of uncertainty reduction is readily apparent. AAE and PCA+HT provide about the same level of accuracy and uncertainty reduction in posterior predictions for pressure. For saturation, however, the AAE posterior results display smaller differences than the PCA+HT posterior results. This is consistent with our previous comment regarding the tendency of PCA+HT to overestimate posterior uncertainty. 

As a final assessment. we calculate the fraction of the true data that falls within specific ranges of the posterior results. This is quantified in terms of coverage probability (CP), calculated as CP=$\frac{N_c}{N_{\text{full}}}$, where $N_c$ is the number of data variables that fall within the specific uncertainty range and $N_{\text{full}}$ is the total number of data variables. For any specified uncertainty range, the target CP value follows directly. For example, 80\% of the data should fall within the P$_{10}$--P$_{90}$ range, 50\% should fall within the P$_{25}$--P$_{75}$ range, etc. 

\begin{figure}[htbp!]
\centering
\includegraphics[width = 0.75\textwidth]{ 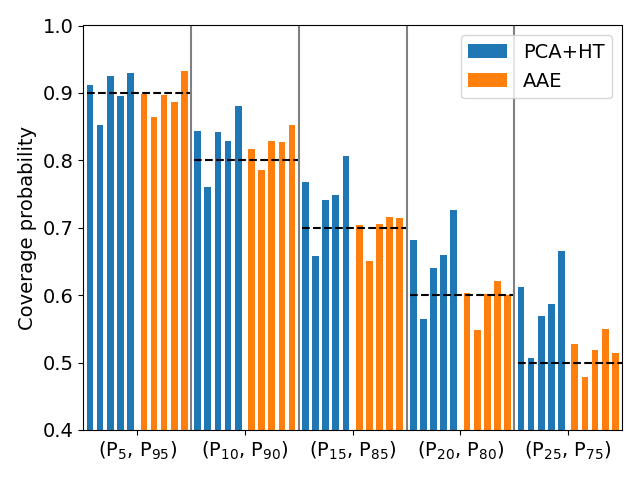}
\caption{Coverage probability (CP) using PCA+HT and AAE for five true models. Target CP values indicated by the dashed lines.} \label{fig:cp}
\end{figure}

CP results are presented in Fig.~\ref{fig:cp}. DSI results are shown for all five true models (each true model is indicated by a vertical bar) using PCA+HT (blue bars) and AAE (orange bars). Five uncertainty ranges are considered, with the target results indicated by the dashed lines. The AAE results are for the most part closer to the target CP values than the PCA+HT results. Consistent with the observations in  \cite{jiang2021data}, we see that PCA+HT tends to overestimate posterior uncertainty. True model~2 is an exception, however, possibly because it lies at the edge of the prior for both average pressure and saturation. In terms of overall CP values, if we average over the five true models, the results for AAE are 0.896, 0.822, 0.698, 0.595 and 0.518 for the five uncertainty ranges considered. For PCA+HT these results are 0.903, 0.831, 0.744, 0.655 and 0.588. Given that the target values are 0.9, 0.8, 0.7, 0.6 and 0.5, these results again suggest that AAE provides more accurate uncertainty quantification than PCA+HT.

\section{Concluding remarks}\label{sec:conclusion}
In this work, we introduced a deep-learning-based spatio-temporal data parameterization procedure and implemented it into DSI. The DSI framework directly provides posterior predictions of pressure and CO$_2$ saturation fields, at multiple time steps, conditioned to observed data. The adversarial autoencoder (AAE) parameterization provides dimension reduction, while retaining the physical characteristics and correlations, for the pressure and saturation fields. A convolutional long short-term memory (convLSTM) network within the AAE captures temporal evolution. Predictions are generated by DSI using the AAE-based parameterization and ESMDA for posterior sampling.

The new DSI framework was applied to models of an industrial-scale carbon storage operation. The geological scenario parameters (i.e., metaparameters characterizing mean and standard deviation of permeability and porosity, correlation lengths, etc.) were considered to be uncertain. This corresponds to a high degree of variability as each set of scenario parameters characterizes an infinite number of geological realizations. A total of 1500 realizations were simulated to provide prior data and to train the AAE parameterization. Accurate reconstruction of pressure and saturation fields and prior flow statistics, compared with reference simulation results, was demonstrated.  

The overall DSI process was then applied to provide posterior predictions for pressure and CO$_2$ saturation conditioned to monitoring well data. Local grid refinement results were used to estimate model resolution error for history matching. Detailed data assimilation results using DSI with AAE were presented for two true models. The DSI procedure required about 4~minutes to generate a full set of posterior results for each case. Uncertainty reduction for a range of pressure and saturation quantities was demonstrated. The impact of total error on posterior results was assessed. Three additional true models were then considered to further evaluate the performance of our new treatments. Comparison of summary results (mean absolute differences and coverage probability) with the simpler PCA-based parameterization demonstrated some advantages of the AAE procedure. 

There are several directions that could be considered in future work. The parameterization and DSI workflow presented here can be extended to treat coupled flow and geomechanics problems. This would allow the method to be used to predict stress fields, and the potential for fault slip due to CO$_2$ injection, from pressure and strain measurements. The development of treatments to incorporate model parameters into the DSI framework would enable a degree of linkage between traditional model-based approaches and DSI. This would be very useful in cases where posterior model parameters are of interest. It is also important to extend the error modeling approach to treat error from a range of sources (e.g., unresolved physics, uncertain multiphase flow or dissolution models) and to correct for model bias. Finally, linkage of the DSI procedure with multiobjective monitoring well optimization would enable decision makers to quantify the costs and tradeoffs of different monitoring strategies.

\section*{CRediT authorship contribution statement}
\textbf{Su Jiang}: Conceptualization, Methodology, Software, Visualization, Formal analysis, Results interpretation, Writing -- original draft. \textbf{Louis J. Durlofsky}: Supervision, Conceptualization, Resources, Formal analysis, Results interpretation, Writing -- review \& editing.

\section*{Declaration of competing interest}
\noindent The authors declare that they have no known competing financial interests or personal relationships that could have appeared to influence the work reported in this paper.

\section*{Acknowledgements}
\noindent We are grateful to the Stanford Smart Fields Consortium for partial funding of this work. We thank Dylan Crain for assistance with the geomodels. We acknowledge the Stanford Center for Computational Earth \& Environmental Science for providing the computational resources used in this work. 

\section*{Code availability}
\noindent The code used in this study will be made available on github when this paper is published. Please contact Su Jiang (sujiang@stanford.edu) for earlier access.

\bibliographystyle{elsarticle-num-names} 
\bibliography{reference}

\end{document}